\documentclass[runningheads]{llncs}
\usepackage{graphicx}

\usepackage{tikz}
\usepackage{comment} 
\usepackage{amsmath,amssymb} 
\usepackage{color}

\usepackage{bbm}
\usepackage{multirow}
\usepackage[justification=centering]{subfig}
\usepackage{booktabs}
\usepackage{cite}
\usepackage{pdfpages}
\usepackage{dblfloatfix}

\usepackage{hyperref}
\hypersetup{
    colorlinks=true,
    urlcolor=cyan,
}

\begin{document}
\pagestyle{headings}
\mainmatter
\def\ECCVSubNumber{2246}  

\title{Classes Matter: A Fine-grained Adversarial Approach to Cross-domain Semantic Segmentation} 

\titlerunning{Classes Matter: A Fine-grained Adversarial Approach}
%
\author{Haoran Wang\textsuperscript{1}* \and
Tong Shen\textsuperscript{2}* \and
Wei Zhang\textsuperscript{2} \and
Ling-Yu Duan\textsuperscript{3} \and
Tao Mei\textsuperscript{2} \\
\quad \\
\small \textsuperscript{1}ETH Z{\"u}rich\quad \textsuperscript{2}JD AI Research\quad \textsuperscript{3}Peking University
}

%
\authorrunning{H. Wang et al.}
%

\institute{}
\maketitle

\def\thefootnote{*}\footnotetext{These authors contributed equally. This work was performed when Haoran Wang was visiting JD AI research as a research intern.}\def\thefootnote{\arabic{footnote}}

\begin{abstract}
Despite great progress in supervised semantic segmentation, a large performance drop is usually observed when deploying the model in the wild. 
Domain adaptation methods tackle the issue by aligning the source domain and the target domain. However, most existing methods attempt to perform the alignment from a holistic view, ignoring the underlying class-level data structure in the target domain.
To fully exploit the supervision in the source domain, we propose a fine-grained adversarial learning strategy for class-level feature alignment while preserving the internal structure of semantics across domains. 
We adopt a fine-grained domain discriminator that not only plays as a domain distinguisher, but also differentiates domains at class level. The traditional binary domain labels are also generalized to domain encodings as the supervision signal to guide the fine-grained feature alignment.
%
%
An analysis with Class Center Distance (CCD) validates that our fine-grained adversarial strategy achieves better class-level alignment compared to other state-of-the-art methods.
%
Our method is easy to implement and its effectiveness is evaluated on three classical domain adaptation tasks, i.e., GTA5$\to$Cityscapes, SYNTHIA$\to$Cityscapes and Cityscapes$\to$Cross-City.
%
Large performance gains show that our method outperforms other global feature alignment based and class-wise alignment based counterparts. The code is publicly available at \url{https://github.com/JDAI-CV/FADA}.

\end{abstract}

\section{Introduction}

\begin{figure}[t]
\centering
   \includegraphics[height=6.5cm]{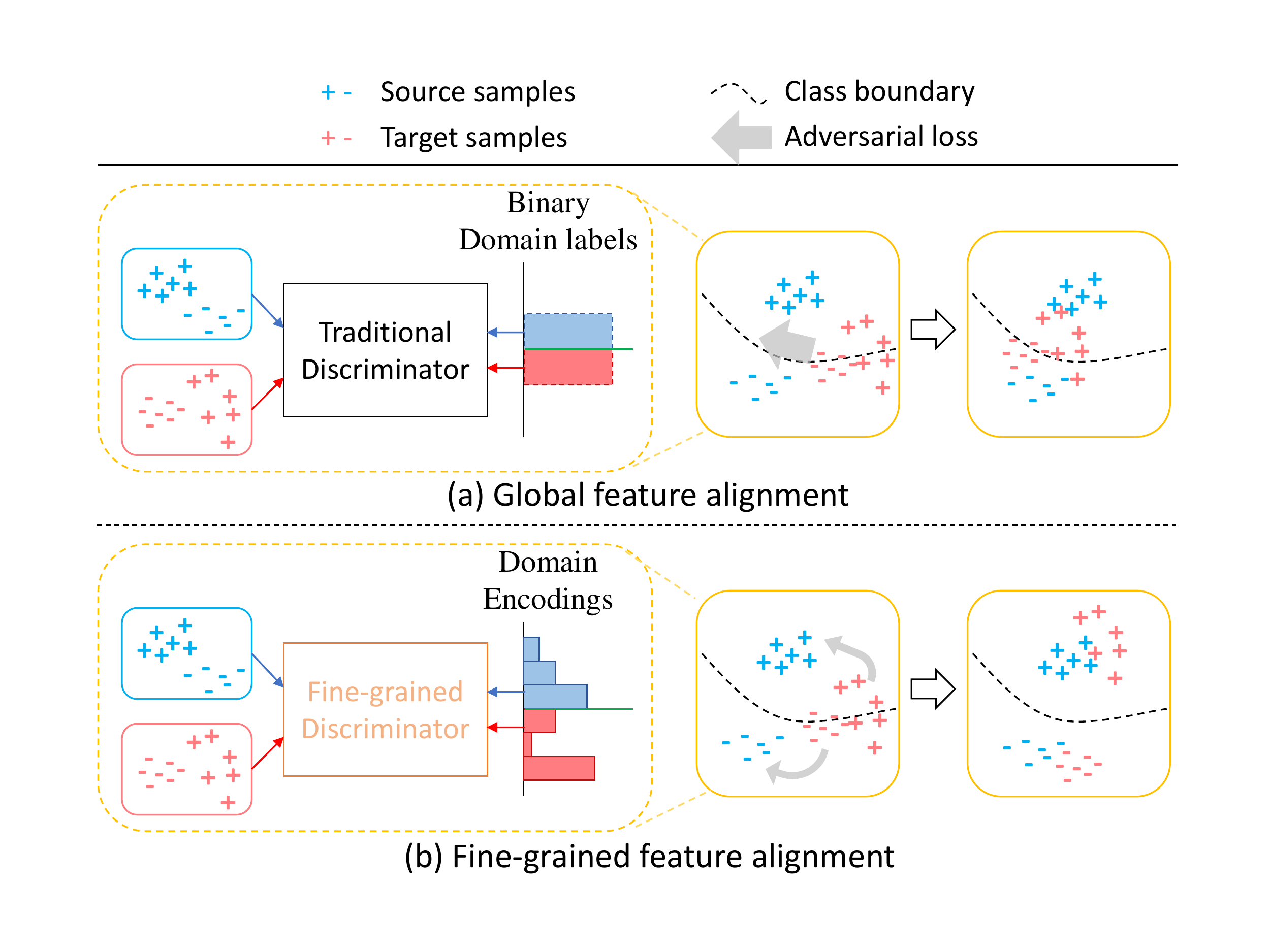}
   \caption{Illustration of traditional and our fine-grained adversarial learning. Traditional adversarial learning pursues the marginal distribution alignment while ignoring the semantic structure inconsistency between domains. We propose to use a fine-grained discriminator to enable class-level alignment.
   }
\label{fig:motivation}
\end{figure}

 The success of semantic segmentation \cite{Shelhamer2017FCN} in recent years is mostly driven by a large amount of accessible labeled data. However, collecting massive densely annotated data for training is usually a labor-intensive task \cite{Cordts2016Cityscapes}. 
 Recent advances in computer graphics provide an alternative for replacing expensive human labor. Through physically based rendering, we can obtain photo-realistic images with the pixel-level ground-truth readily available in an effortless way \cite{Richter_2016_ECCV_gtav,Ros_2016_CVPR_synthia}.


However, performance drop is observed when the model trained with synthetic data (a source domain) is applied in realistic scenarios (a target domain), because the data from different domains usually share different distributions. This phenomenon is known as domain shift problem \cite{covariate_shift}, which poses a challenge to cross-domain tasks \cite{hoffman2016fcns}.

Domain adaptation aims to alleviate the domain shift problem by aligning the feature distributions of the source and the target domain. 
A group of works focus on adopting an adversarial framework, where a domain discriminator is trained to distinguish the target samples from the source ones, while the feature network tries to fool the discriminator by generating domain-invariant features~\cite{hoffman2016fcns,Tsai_adaptseg_2018,Hoffman_cycada2017,zou2018cbst,Yawei2019Taking,saito2017maximum,Zhang_2017_ICCV_CDA, chen2018_dafrcnn, fcan}.


Although impressive progress has been achieved in domain adaptive semantic segmentation, most of prior works strive to align global feature distributions without paying much attention to the underlying structures among classes. However, as discussed in recent works \cite{NIPS2018_co_regularized, Chen_2019_CVPR}, matching the marginal feature distributions does not guarantee small expected error on the target domain. The class conditional distributions should also be aligned, meaning that class-level alignment also plays an important role. As illustrated in Figure \ref{fig:motivation}, the upper part shows the result of global feature alignment where the two domains are well-aligned but some samples are falsely mixed up. This motivates us to incorporate class information into the adversarial framework to enable fine-grained feature alignment. As illustrated in the bottom of Figure \ref{fig:motivation}, features are expected to 
be aligned according to specific classes.


There have been some pioneering works \cite{Yawei2019Taking,no_more_discrimination} trying to address this problem. Chen et al. \cite{no_more_discrimination} propose to use several independent discriminators to perform class-wise alignment, but independent discriminators might fail to capture the relationships between classes. Luo et al. \cite{Yawei2019Taking} introduce an self-adaptive adversarial loss to apply different weights to each region. However, in fact, they do not explicitly incorporate class information in their methods, which might fail to promote class-level alignment.


Our motivation is to directly incorporate class information into the discriminator and encourage it to align features at a fine-grained level. Traditional adversarial training has been proven effective for aligning features by using a binary domain discriminator to model distribution $P(d|f)$ ($d$ refers to domain and $f$ is the feature extracted from input data). By confusing such a discriminator, expecting $P(d=0|f)\approx P(d=1|f)$ where 0 stands for the source domain and 1 for the target domain, the features become domain invariant and well aligned. To further take classes into account, we split the output into multiple channels according to $P(d|f)=\sum_{c=1}^{K}P(d,c|f)$ (where $c$ refers to classes $\{1,\dots,K\}$). We directly model the discriminator as $P(d,c|f)$ to formulate a fine-grained domain alignment task. Although in the setting of domain adaptation the category-level labels for target domain are inaccessible, we find that the model predictions on target domain also contain class information and prove that it is possible to supervise the discriminator with the predictions on both domains. In the adversarial learning process, class information is incorporated and the features are expected to be aligned according to specific classes.

In this paper, we propose such a fine-grained adversarial learning framework for domain adaptive semantic segmentation (FADA). As illustrated in Figure \ref{fig:motivation}, we represent the supervision of traditional discriminator at a fine-grained semantic level, which enables our fine-grained discriminator to capture rich class-level information. The adversarial learning process is performed at fine-grained level, so the features are expected to be adaptively aligned according to their corresponding semantic categories. The class mismatch problem, which broadly exists in the global feature alignment, is expected to be further suppressed. Correspondingly, by incorporating class information, the binary domain labels are also generalized to a more complex form, called ``domain encodings'' to serve as the new supervision signal. Domain encodings could be extracted from the network's predictions on both domains. Different strategies of constructing domain encodings will be discussed. We conduct an analysis with Class Center Distance to demonstrate the effectiveness of our method regarding class-level alignment. Our method is also evaluated on three popular cross-domain benchmarks and presents new state-of-the-art results.


The main contributions of this paper are summarized below.
\begin{itemize}
  \item We propose a fine-grained adversarial learning framework for cross-domain semantic segmentation that explicitly incorporates class-level information.
  
  \item The fine-grained learning framework enables class-level feature alignment, which is further verified by analysis using Class Center Distance. 
  
  \item We evaluate our methods with comprehensive experiments. Significant improvements compared to other state-of-the-art methods are achieved on popular domain adaptive segmentation tasks including GTA5 $\to$ Cityscapes, SYNTHIA $\to$ Cityscapes and Cityscapes $\to$ Cross-City.
\end{itemize}


\section{Related Work}

\subsection{Semantic Segmentation}
Semantic segmentation is a task of predicting unique semantic label for each pixel of the input image. With the advent of deep convolutional neural networks, the academia of computer vision witnesses a huge progress in this field. FCN \cite{Shelhamer2017FCN} triggered the interests in introducing deep learning for this task. Many follow-up methods are proposed to enlarge the receptive fields to cover more context information \cite{deeplabv1, deeplabv2, zhao2017pspnet, deeplabv3plus2018}. Among all these works, the family of Deeplab \cite{deeplabv1, deeplabv2, deeplabv3plus2018} attracts a lot of attention and has been widely applied in many works for their simplicity and effectiveness. 

\subsection{Domain Adaptation}
Domain adaptation strives to address the performance drop caused by the different distributions of training data and testing data. In the recent years, several works are proposed to approach this problem in image classification \cite{saito2017maximum,Chen_2019_CVPR}. Inspired by the theoretical upper bound of risk in target domain \cite{Ben-David2010}, some pioneering works suggest to optimize some distance measurements between 
the two domains to align the features \cite{mmd,deep_coral}. Recently, motivated by GAN \cite{Goodfellow:2014:GAN:2969033.2969125}, adversarial training becomes popular for its power to align features globally \cite{Tsai_adaptseg_2018,no_more_discrimination,saito2017maximum}.

\subsection{Domain Adaptive Semantic Segmentation}
Unlike domain adaptation for image classification task, domain adaptive semantic segmentation receives less attention for its difficulty even though it supports many important applications including autonomous driving in the wild \cite{hoffman2016fcns, chen2018_dafrcnn}.
Based on the theoretical insight \cite{Ben-David2010} on domain adaptive classification, most works follow the path of shortening the domain discrepancy between the two domains.  Large progress is achieved through optimization by adversarial training or explicit domain discrepancy measures \cite{Tsai_adaptseg_2018, Hoffman_cycada2017,hoffman2016fcns}. In the context of domain adaptive semantic segmentation task, AdaptSegnet \cite{Tsai_adaptseg_2018} attempts to align the distribution in the output space. Inspired by CycleGAN \cite{CycleGAN2017}, CyCADA \cite{Hoffman_cycada2017} suggests to adapt the representation in pixel-level and feature-level. There are also many works focusing on aligning different properties between two domains such as entropy \cite{vu2018advent} and information \cite{Luo_2019_ICCV}. 

Although huge progress has been made in this field, most of existing methods share a common limitation: Enforcing global feature alignment would inevitably mix samples with different semantic labels together when drawing two domains closer, which usually results in a mismatch of classes from different domains. CLAN \cite{Yawei2019Taking} is a pioneer work to address category-level alignment. It suggests applying different adversarial weight to different regions, but it does not directly and explicitly incorporate the classes into the model.


\section{Method}
\begin{figure}[t]
\centering
\includegraphics[width=0.8\textwidth]{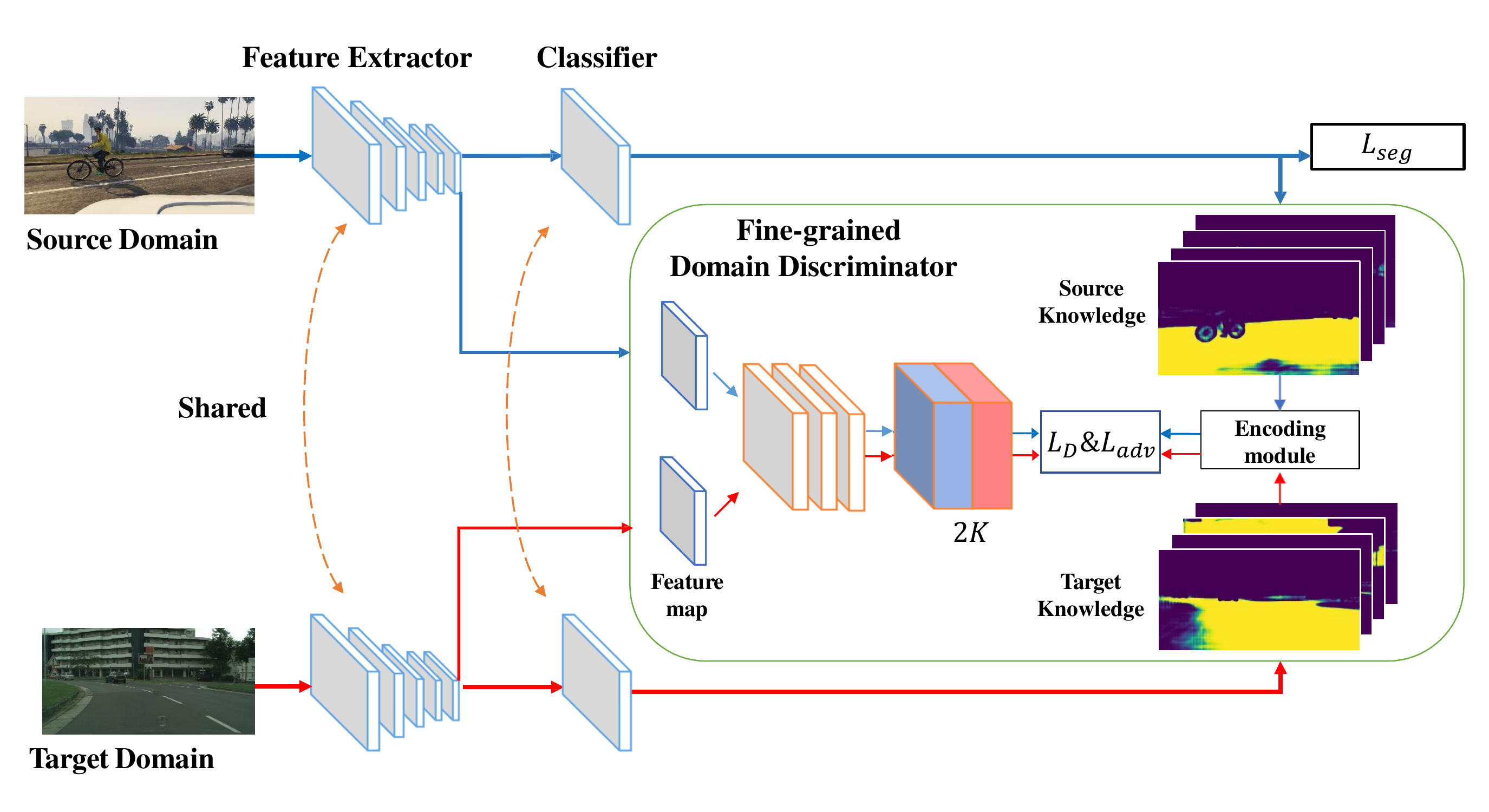}
   \caption{Overview of the proposed fine-grained adversarial framework. Images from the source domain and target domain are randomly picked and fed to the feature extractor and the classifier. A segmentation loss is computed with the source predictions and the source annotations to help the segmentation network to generate discriminative features and learn task specific knowledge. The semantic features from both domains are fed to the convolutional fine-grained domain discriminator. The discriminator strives to distinguish the feature's domain information at a fine-grained class level using the domain encodings processed from the sample predictions.}
   
\label{fig:architecture}
\end{figure}

\subsection{Revisit Traditional Feature Alignment}
Semantic segmentation aims to predict per-pixel unique label for the input image \cite{Shelhamer2017FCN}. 
In an unsupervised domain adaptation setting for semantic segmentation, we have access to a collection of labeled data $X_\mathcal{S}=\{(x_i^{(s)}, y_i^{(s)})\}_{i=1}^{n_s}$ in a source domain $\mathcal{S}$, and unlabeled data $X_\mathcal{T}=\{x_j^{(t)}\}_{j=1}^{n_t}$ in a target domain $\mathcal{T}$ where $n_s$ and $n_t$ are the numbers of samples from different domains. Domain $\mathcal{S}$ and domain $\mathcal{T}$ share the same $K$ semantic class labels $\{1,\dots,K\}$.
The goal is to learn a segmentation model $G$ which could achieve a low expected risk on the target domain. 
Generally, segmentation network $G$ could be divided into a feature extractor $F$ and a multi-class classifier $C$, where $G=C\circ F$. 

Traditional feature-level adversarial training relies on a binary domain discriminator $D$ to align the features extracted by $F$ on both domains. Domain adaptation is tackled by alternatively optimizing $G$ and $D$ with two steps:

(1) $D$ is trained to distinguish features from different domains. This process is usually achieved by fixing $F$ and $C$ and solving:


\begin{equation}
\label{loss_bin_D}
	\min_D \mathcal{L}_{D} = -\sum_{i=1}^{n_s} (1-d)\log{P(d=0|f_i)} 
	-\sum_{j=1}^{n_t}d\log{P(d=1|f_j)}
\end{equation}
where $f_i$ and $f_j$ are the features extracted by $F$ on source sample $x_i^{(s)}$ and target sample $x_j^{(t)}$; $d$ refers to the domain variable where 0 refers to the source domain and 1 refers to the target domain. $P(d|f)$ is the probability output from the discriminator.


(2) $G$ is trained with the task loss $\mathcal{L}_{seg}$ on the source domain and the adversarial loss $\mathcal{L}_{adv}$ on the target domain. This process requires fixing $D$ and updating $F$ and $C$:


\begin{equation}
    \min_{F, C}\mathcal{L}_{seg}+\lambda_{adv}\mathcal{L}_{adv}
\end{equation}

The cross-entropy loss $\mathcal{L}_{seg}$ on source domain minimizes the difference between the prediction and the ground truth, which helps $G$ to learn the task specific knowledge.
\begin{equation}
\label{loss_seg}
    \mathcal{L}_{seg}=-\sum_{i=1}^{n_s}\sum_{k=1}^{K}y_{ik}^{(s)}\log p_{ik}^{(s)},
\end{equation}
where $p_{ik}^{(s)}$ is the probability confidence of source sample $x_i^{(s)}$ belonging to semantic class k predicted by $C$, $y_{ik}^{(s)}$ is the entry for the one-hot label.

The adversarial loss $\mathcal{L}_{adv}$ is used to confuse the discriminator to encourage $F$ to generate domain invariant features.

\begin{equation}
\begin{split}\label{loss_bin_adv}
 \mathcal{L}_{adv} = -\sum_{j=1}^{n_t}\log{P(d=0|f_j)}
\end{split}
\end{equation}

\subsection{Fine-grained Adversarial Learning}
To incorporate the class information into the adversarial learning framework, we propose a novel discriminator and enable a fine-grained adversarial learning process. The whole pipeline is illustrated in Figure \ref{fig:architecture}.

The traditional adversarial training strives to align the marginal distribution by confusing a binary discriminator. To make the discriminator not merely focus on distinguishing domains, we split each of the two output channels of the binary discriminator into K channels and encourage a fine-grained level adversarial learning. With this design, the predicted confidence for domains is represented as a confidence distribution over different classes, which enables the new fine-grained discriminator to model more complex underlying structures between classes, thus encouraging class-level alignment.


Correspondingly, the binary domain labels are also converted to a general form, namely domain encodings, to incorporate class information. Traditionally, the domain labels used for training the binary discriminator are $[1,0]$ and $[0,1]$ for the source and target domains respectively. The domain encodings are represented as a vector $[\mathbf{a};\mathbf{0}]$ and $[\mathbf{0};\mathbf{a}]$ for the two domains respectively, where $\mathbf{a}$ is the knowledge extracted from the classifier C represented by a $K$-dimensional vector; $\mathbf{0}$ is an all-zero $K$-dimensional vector. The choices of how to generate domain knowledge $\mathbf{a}$ will be discussed in Section \ref{class_level_info}.

During the training process, the discriminator not only tries to distinguish domains, but also learns to model class structures. The $\mathcal{L}_D$ in Equation \ref{loss_bin_D} becomes: 


\begin{equation}
\begin{split}\label{loss_D}
\mathcal{L}_{D} = &-\sum_{i=1}^{n_s}\sum_{k=1}^{K} a_{ik}^{(s)}\log{P(d=0, c=k|f_i)} \\
&-\sum_{j=1}^{n_t}\sum_{k=1}^{K} a_{jk}^{(t)}\log{P(d=1, c=k|f_j)}
\end{split}
\end{equation}

where $a_{ik}^{(s)}$ and $a_{jk}^{(t)}$ are the $k$th entries of the class knowledge for the source sample $i$ and target sample $j$. The adversarial loss $\mathcal{L}_{adv}$ used to confuse the discriminator and guide the generation of domain-invariant features in Equation \ref{loss_bin_adv}  becomes:


\begin{equation}
\label{loss_adv}
    \mathcal{L}_{adv}=-\sum_{j=1}^{n_t}\sum_{k=1}^{K}a_{jk}^{(t)}\log{P(d=0, c=k|f_j)},
\end{equation}

$\mathcal{L}_{adv}$ is designed to maximize the probability of features from target domain being considered as the source features without hurting the relationship between features and classes.



The overall network in Figure \ref{fig:architecture} is used in the training stage. During inference, the domain adaptation component is removed and one only needs to use the original segmentation network with the adapted weights. 

\begin{figure}[t]
\centering
   \includegraphics[width=0.8\textwidth]{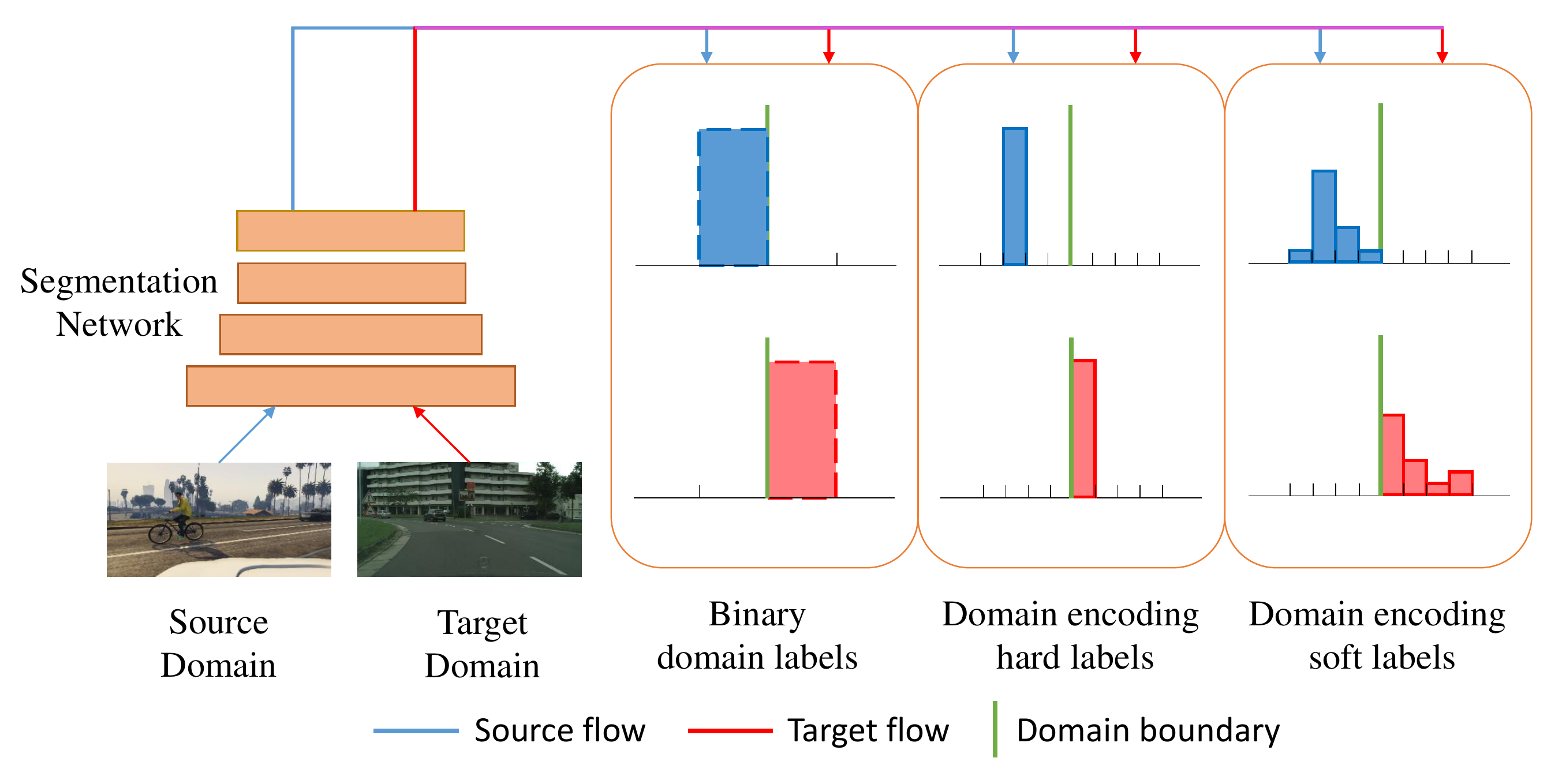}
   \caption{Illustration of different strategies to generate domain encodings. Here we compare three different strategies to extract knowledge from segmentation network for constructing domain encodings: binary domain labels, one-hot hard labels and multi channel soft labels.}
\label{fig:strategies}
\end{figure}

\subsection{Extracting class knowledge for domain encodings}
\label{class_level_info}
Now that we have a fine-grained domain discriminator, which could adaptively align features according to the class-level information contained in domain encodings, another challenge raises: how to get the class knowledge $a_{ik}^{(s)}$ and $a_{ik}^{(t)}$ in Equations \ref{loss_D} and \ref{loss_adv} to construct domain encoding for each sample? Considering that in the unsupervised domain adaptive semantic segmentation task none of annotations in target domain is accessible, it seems contradictory to use the class knowledge on the target domain for guiding class-level alignment. However, during training, with ground-truth annotations from the source domain, the classifier $C$ learns to map features into the semantic classes. Considering that the source domain and the target domain share the same semantic classes, it would be a natural choice to use the predictions of $C$ as knowledge to supervise the discriminator.

As illustrated in equations \ref{loss_D} and \ref{loss_adv}, the class knowledge for optimizing the fine-grained discriminator works as the supervision signal. The choices of $a_{ik}^{(s)}$ and $a_{jk}^{(t)}$ are open to many possibilities. For specific tasks, people could design different forms to produce class knowledge with prior knowledge. Here we discuss two general solutions to extract class knowledge from network predictions for constructing domain encodings. Because the class-level knowledge for different domains could be extracted in the same way, in the following discussion we would use $a_k$ to represent $k$th entry for a single sample without differentiating the domain.


The one-hot hard labels could be a straightforward solution for generating knowledge, which could be denoted as: 

\begin{equation}
        a_{k}=
        \begin{cases}
        \begin{array}{lll}
1&\mbox{if $k=\arg\max\limits_{k}p_k$}\\
0 &\mbox{otherwise}
        \end{array}
        \end{cases}
\end{equation}
where $p_{k}$ is the softmax probability output of $C$ for class $k$.
In this way, only the most confident class is selected. In practice, in order to remove the impact of noisy samples, we can select samples whose confidence is higher than a certain threshold and ignore those with low confidence.

Another alternative is multi-channel soft labels, which has the following definition: 
\begin{equation}
    a_{k}=\frac{\exp{(\frac{z_k}{T})}}{\sum_{j=1}^K\exp{(\frac{z_j}{T})}}
\end{equation}
where $z_k$ is $k$th entry of logits and $T$ is a temperature to encourage soft probability distribution over classes. Note that during training, an additional regularization could also be applied. For example, we practically find that clipping the values of the soft labels by a given threshold achieves more stable performance because it prevents from overfitting to certain classes.
  
An illustrative comparison of these two strategies with the traditional binary domain labels is presented in Figure \ref{fig:strategies}. We also conduct experiments in section \ref{vs} to demonstrate the performance of different strategies.

\section{Experiments}
\begin{table}[t]
    \caption{Experimental results for Cityscapes $\to$ Cross-City.}
    \centering
    \resizebox{\textwidth}{!}{
    \renewcommand{\arraystretch}{1.2}
    \begin{tabular}{l | l | c c c c c c c c c c c c c | c}
    \hline
    \multicolumn{16}{c}{\textbf{Cityscapes} $\to$ \textbf{Cross-City}} \\
    \hline
    City & Method & road & sidewalk & building & light & sign & veg & sky & person & rider & car & bus & mbike & bike & mIoU\\
    \hline
    \multirow{5}{*}{Rome} 
    & Source Dilation-Frontend & 77.7& 21.9& 83.5& 0.1& 10.7& 78.9& 88.1& 21.6& 10.0& 67.2& 30.4& 6.1& 0.6& 38.2\\
    & Cross-City \cite{no_more_discrimination} & 79.5 & 29.3 & 84.5 & 0.0 & 22.2 & 80.6 & 82.8 & 29.5 & 13.0 & 71.7 & 37.5 & 25.9 & 1.0 & 42.9 \\
    \cline{2-16}
    & Source DeepLab-v2 & 83.9& 34.3& 87.7& 13.0& 41.9 &84.6& 92.5& 37.7& 22.4& 80.8& 38.1& 39.1& 5.3& 50.9\\
    & AdaptSegNet \cite{Tsai_adaptseg_2018}& 83.9 & 34.2 &\textbf{88.3} & 18.8 & \textbf{40.2} & \textbf{86.2} & \textbf{93.1} & 47.8 & 21.7 & 80.9 & \textbf{47.8} & 48.3 & 8.6 & 53.8\\
    & FADA & \textbf{84.9} & \textbf{35.8} &\textbf{88.3} & \textbf{20.5} & 40.1 & 85.9 & 92.8 &\textbf{56.2} & \textbf{23.2} & \textbf{83.6} & 31.8 & \textbf{53.2} & \textbf{14.6} &\textbf{54.7} \\
    \hline
    
    \multirow{5}{*}{Rio} & Source Dilation-Frontend & 69.0& 31.8& 77.0& 4.7& 3.7& 71.8& 80.8& 38.2& 8.0& 61.2& 38.9& 11.5& 3.4 &38.5\\
    & Cross-City \cite{no_more_discrimination} & 74.2 & 43.9 & 79.0 & 2.4 & 7.5 & 77.8 & 69.5 & 39.3 & 10.3 & 67.9 & \textbf{41.2} & 27.9 & 10.9 & 42.5 \\
    \cline{2-16}
    & Source DeepLab-v2 & 76.6 &47.3 &82.5& \textbf{12.6}& 22.5& 77.9& 86.5& 43.0& 19.8& 74.5& 36.8& 29.4& 16.7& 48.2\\
    & AdaptSegNet \cite{Tsai_adaptseg_2018}& 76.2 & 44.7 & \textbf{84.6} & 9.3 & \textbf{25.5} & \textbf{81.8} & 87.3 & 55.3 & \textbf{32.7} & 74.3 & 28.9 & 43.0 & 27.6 & 51.6 \\
    & FADA & \textbf{80.6} & \textbf{53.4} & 84.2 & 5.8 &23.0 & 78.4& \textbf{87.7} & \textbf{60.2} & 26.4 & \textbf{77.1} & 37.6 & \textbf{53.7} & \textbf{42.3} & \textbf{54.7}\\
    \hline
    
    \multirow{5}{*}{Tokyo} 
    & Source Dilation-Frontend & 81.2& 26.7& 71.7& 8.7 &5.6& 73.2& 75.7& 39.3& 14.9& 57.6& 19.0& 1.6& 33.8& 39.2\\
    & Cross-City \cite{no_more_discrimination} & 83.4 & 35.4 & 72.8 & 12.3 & 12.7 & 77.4 & 64.3 & 42.7 & 21.5 & 64.1 & \textbf{20.8} & 8.9 & 40.3 & 42.8 \\
    \cline{2-16}
    & Source DeepLab-v2 & 83.4& 35.4& 72.8& 12.3& 12.7& 77.4& 64.3& 42.7& 21.5& 64.1& \textbf{20.8}& 8.9& 40.3& 42.8\\
    & AdaptSegNet \cite{Tsai_adaptseg_2018}& 81.5 & 26.0 & 77.8 & \textbf{17.8} & \textbf{26.8} & 82.7 & 90.9 & 55.8 & \textbf{38.0} & \textbf{72.1} & 4.2 & 24.5 & 50.8 & 49.9\\
    & FADA & \textbf{85.8} & \textbf{39.5} & \textbf{76.0} & 14.7 & 24.9 & \textbf{84.6} & \textbf{91.7} & \textbf{62.2} & 27.7 & 71.4 & 3.0 & \textbf{29.3} & \textbf{56.3} &\textbf{51.3} \\
    \hline
    
    \multirow{5}{*}{Taipei} 
    &Source Dilation-Frontend & 77.2& 20.9& 76.0& 5.9& 4.3& 60.3& 81.4& 10.9& 11.0& 54.9& 32.6& 15.3& 5.2& 35.1\\
    & Cross-City \cite{no_more_discrimination} & 78.6& 28.6& 80.0& 13.1& 7.6& 68.2& 82.1& 16.8& 9.4& 60.4& 34.0& 26.5& 9.9& 39.6 \\
    \cline{2-16}
    & Source DeepLab-V2 & 78.6& 28.6& 80.0& 13.1& 7.6 &68.2 &82.1 &16.8 &9.4 &60.4 &34.0 &26.5 &9.9& 39.6\\
    & AdaptSegNet \cite{Tsai_adaptseg_2018} & 81.7& 29.5& 85.2& \textbf{26.4}& 15.6& 76.7& 91.7& 31.0& 12.5& 71.5& \textbf{41.1}& 47.3& 27.7& 49.1\\
    & FADA &\textbf{86.0} & \textbf{42.3} & \textbf{86.1} & 6.2 & \textbf{20.5} & \textbf{78.3} & \textbf{92.7} & \textbf{47.2} & \textbf{17.7} &\textbf{72.2} & 37.2 & \textbf{54.3} & \textbf{44.0}&\textbf{52.7}\\
    \hline
    \end{tabular}}
    
    \label{tab:crosscity_table}
\end{table}

\subsection{Datasets}
We present a comprehensive evaluation of our proposed method on three popular unsupervised domain adaptive semantic segmentation benchmarks, e.g., Cityscapes $\to$ Cross-City, SYNTHIA $\to$ Cityscapes, and GTA5 $\to$ Cityscapes.

\noindent\textbf{Cityscapes}\quad Cityscapes \cite{Cordts2016Cityscapes} is a real-world urban scene dataset consisting of a training set with 2,975 images, a validation set with 500 images and a testing set with 1,525 images. Following the standard protocols \cite{hoffman2016fcns, Hoffman_cycada2017, Tsai_adaptseg_2018}, we use the 2,975 images from Cityscapes training set as the unlabeled target domain training set and evaluate our adapted model on the 500 images from the validation set. 

\noindent\textbf{Cross-City}\quad Cross-City \cite{no_more_discrimination} is an urban scene dataset collected with Google Street View. It contains 3,200 unlabeled images and 100 annotated images of four different cities respectively. The annotations of Cross-City share 13 classes with Cityscapes.

\noindent\textbf{SYNTHIA}\quad SYNTHIA \cite{Ros_2016_CVPR_synthia} is a synthetic urban scene dataset. We pick its subset SYNTHIA-RAND-CITYSCAPES, which shares 16 semantic classes with Cityscapes, as the source domain. In total, 9,400 images from SYNTHIA dataset are used as source domain training data for the task.

\noindent\textbf{GTA5}\quad GTA5 dataset \cite{Richter_2016_ECCV_gtav} is another synthetic dataset sharing 19 semantic classes with Cityscapes. 24,966 urban scene images are collected from a physically-based rendered video game Grand Theft Auto V (GTAV) and are used as source training data. 

\begin{table}[t]
\caption{Experimental results for SYNTHIA $\to$ Cityscapes.}
\begin{center}

\resizebox{\textwidth}{!}{
    \renewcommand{\arraystretch}{1.2}
    \begin{tabular}{ c| l | c c c c c c c c c c c c c c c c | c | c}
 \hline
 \multicolumn{20}{c}{\textbf{SYNTHIA} $\to$ \textbf{Cityscapes}} \\
 \hline
Backbone & Method & Road & SW & Build & Wall & Fence & Pole & TL & TS & Veg. & Sky & PR & Rider & Car & Bus & Motor & Bike & mIoU & mIoU*\\
 \hline

\multirow{12}{*}{VGG-16}
 & FCNs in the wild \cite{hoffman2016fcns}& 11.5 &19.6& 30.8& 4.4& 0.0& 20.3& 0.1& 11.7& 42.3& 68.7& 51.2&3.8& 54.0& 3.2& 0.2& 0.6& 20.2&22.9\\
 &CDA \cite{Zhang_2017_ICCV_CDA} & 65.2 & 26.1& 74.9& 0.1& \textbf{0.5}& 10.7& 3.5& 3.0& 76.1& 70.6& 47.1& 8.2& 43.2& 20.7& 0.7& 13.1& 29.0&34.8\\
 &ST \cite{zou2018cbst} & 0.2&14.5&53.8&1.6&0.0&18.9&0.9&7.8&72.2&80.3&48.1&6.3&67.7&4.7&0.2&4.5&23.9&27.8\\
 &CBST \cite{zou2018cbst} & 69.6&28.7&69.5&\textbf{12.1}&0.1&25.4&\textbf{11.9}&13.6&\textbf{82.0}&81.9&\textbf{49.1}&14.5&66.0&6.6&3.7&\textbf{32.4}&35.4&36.1\\
 &AdaptSegNet \cite{Tsai_adaptseg_2018} & 78.9 &29.2& 75.5&-&-&-& 0.1& 4.8& 72.6& 76.7& 43.4& 8.8& 71.1& 16.0& 3.6& 8.4&-& 37.6\\
 &SIBAN \cite{Luo_2019_ICCV} &70.1&25.7&80.9&-&-&-&3.8&7.2&72.3&80.5&43.3&5.0&73.3&16.0&1.7&3.6&-&37.2\\
 &CLAN \cite{Yawei2019Taking} &80.4 &30.7 &74.7&-&-&-& 1.4& 8.0& 77.1& 79.0& 46.5& 8.9& 73.8& 18.2& 2.2& 9.9&-& 39.3\\
 & AdaptPatch \cite{Tsai_DA4Seg_ICCV19} &72.6 &29.5& 77.2& 3.5& 0.4& 21.0& 1.4& 7.9& 73.3& 79.0& 45.7& 14.5& 69.4& 19.6& 7.4& 16.5& 33.7& 39.6 \\
 &ADVENT \cite{vu2018advent} &67.9 &29.4& 71.9& 6.3& 0.3& 19.9& 0.6& 2.6& 74.9& 74.9& 35.4& 9.6& 67.8& 21.4& 4.1& 15.5& 31.4& 36.6\\

 \cline{2-20}
 &Source only & 10.0&	14.7&	52.4&	4.2&	0.1&	20.9&	3.5&	6.5&	74.3&	77.5&	44.9&	4.9&	64.0&	21.6&	4.2&	6.4&	25.6&	29.6\\
 &Baseline (feat. only)  \cite{Tsai_adaptseg_2018} & 63.6 & 26.8 & 67.3 & 3.8 & 0.3 & 21.5 & 1.0 & 7.4 & 76.1 & 76.5 & 40.5 & 11.2 & 62.1 & 19.4 & 5.3 & 13.2 & 31.0 & 36.2\\
 &FADA & \textbf{80.4}&	\textbf{35.9}&	\textbf{80.9}&	2.5&	0.3&	\textbf{30.4}&	7.9&	\textbf{22.3}&	81.8&	\textbf{83.6}&	48.9&	\textbf{16.8}&	\textbf{77.7}&	\textbf{31.1}&	\textbf{13.5}&	17.9&	\textbf{39.5}&	\textbf{46.0}
\\
 \hline
 \multirow{8}{*}{ResNet-101} 
&SIBAN \cite{Luo_2019_ICCV}&82.5& 24.0& 79.4&-&-&-& 16.5& 12.7& 79.2& 82.8& \textbf{58.3}& 18.0& 79.3& 25.3& 17.6& 25.9 &-&46.3\\
&AdaptSegNet \cite{Tsai_adaptseg_2018} & 84.3 &42.7 &77.5&-&-&-& 4.7& 7.0& 77.9& 82.5& 54.3& 21.0& 72.3& 32.2& 18.9& 32.3&-& 46.7\\
&CLAN \cite{Yawei2019Taking}&81.3 &37.0 &80.1&-&-&-& 16.1& 13.7& 78.2& 81.5& 53.4& 21.2& 73.0& 32.9& 22.6& 30.7&-& 47.8\\
& AdaptPatch \cite{Tsai_DA4Seg_ICCV19} &82.4& 38.0& 78.6& 8.7& \textbf{0.6}& 26.0& 3.9& 11.1& 75.5& \textbf{84.6}& 53.5& 21.6& 71.4& 32.6& 19.3& 31.7& 40.0& 46.5\\
&ADVENT\cite{vu2018advent}&\textbf{85.6} &\textbf{42.2}& 79.7 &8.7 &0.4 &25.9& 5.4& 8.1& 80.4& 84.1& 57.9& \textbf{23.8}& 73.3& 36.4& 14.2& \textbf{33.0}& 41.2& 48.0\\
\cline{2-20}
& Source only& 55.6& 23.8& 74.6& 9.2& 0.2& 24.4& 6.1& 12.1& 74.8& 79.0& 55.3& 19.1& 39.6& 23.3& 13.7& 25.0& 33.5& 38.6\\
& Baseline (feat. only) \cite{Tsai_adaptseg_2018}& 62.4 & 21.9 & 76.3 & \textbf{11.5} & 0.1 & 24.9 & 11.7 & 11.4 & 75.3 & 80.9 & 53.7 & 18.5 & 59.7 & 13.7 & 20.6 & 24.0 & 35.4 & 40.8\\
& FADA& 84.5&40.1&\textbf{83.1}&4.8&0.0&\textbf{34.3}&\textbf{20.1}&\textbf{27.2}&\textbf{84.8}&84.0&53.5&22.6&\textbf{85.4}&\textbf{43.7}&\textbf{26.8}&27.8&\textbf{45.2}&\textbf{52.5}
\\

 \hline
\end{tabular}}
\end{center}

\label{tab:synthia_results}
\end{table}

\begin{table}[t]
\caption{Experimental results for GTA5 $\to$ Cityscapes.}
\begin{center}
\resizebox{\textwidth}{!}{
\renewcommand{\arraystretch}{1.2}
    \begin{tabular}{ c| l | c c c c c c c c c c c c c c c c c c c | c }
 \hline
 \multicolumn{22}{c}{\textbf{GTA5} $\to$ \textbf{Cityscapes}} \\
 \hline
Backbone & Method & Road & SW & Build & Wall & Fence & Pole & TL & TS & Veg. & Terrain & Sky & PR & Rider & Car & Truck & Bus & Train & Motor & Bike & mIoU\\
 \hline

\multirow{13}{*}{VGG-16}
 & FCNs in the wild \cite{hoffman2016fcns}& 70.4 &32.4& 62.1& 14.9& 5.4& 10.9& 14.2& 2.7& 79.2& 21.3& 64.6& 44.1& 4.2& 70.4& 8.0& 7.3& 0.0& 3.5& 0.0& 27.1\\
 &CDA \cite{Zhang_2017_ICCV_CDA} & 74.9 & 22.0& 71.7& 6.0& 11.9& 8.4& 16.3& 11.1& 75.7& 13.3& 66.5& 38.0& 9.3& 55.2& 18.8& 18.9& 0.0& 16.8& 14.6& 28.9\\
 &ST \cite{zou2018cbst} & 83.8 & 17.4 & 72.1 & 14.6 & 2.9 & 16.5 & 16.0 & 6.8 & 81.4 & 24.2 & 47.2 & 40.7 & 7.6 & 71.7 & 10.2 & 7.6 & 0.5 & 11.1 & 0.9&28.1\\
 &CBST \cite{zou2018cbst} & 90.4&50.8&72.0&18.3& 9.5& 27.2& \textbf{28.6} & 14.1& 82.4& 25.1& 70.8& 42.6& 14.5& 76.9& 5.9& 12.5& 1.2& 14.0& \textbf{28.6}& 36.1\\
 &CyCADA \cite{Hoffman_cycada2017} &85.2 & 37.2& 76.5& 21.8& 15.0& 23.8& 22.9& 21.5& 80.5& 31.3& 60.7& 50.5& 9.0& 76.9& 17.1& 28.2& 4.5& 9.8& 0.0& 35.4\\
 &AdaptSegNet \cite{Tsai_adaptseg_2018} & 87.3&29.8& 78.6& 21.1& 18.2& 22.5& 21.5& 11.0& 79.7& 29.6& 71.3& 46.8& 6.5& 80.1& 23.0& 26.9& 0.0& 10.6& 0.3& 35.0\\
 &SIBAN \cite{Luo_2019_ICCV}&83.4 &13.0 &77.8& 20.4& 17.5& 24.6 &22.8& 9.6& 81.3& 29.6& 77.3& 42.7& 10.9& 76.0 &22.8& 17.9& 5.7& 14.2& 2.0 &34.2\\
 &CLAN \cite{Yawei2019Taking}&88.0 &30.6& 79.2& 23.4& 20.5& 26.1& 23.0& 14.8& 81.6& \textbf{34.5}& 72.0& 45.8& 7.9& 80.5& \textbf{26.6}& 29.9& 0.0& 10.7& 0.0& 36.6\\
 & AdaptPatch \cite{Tsai_DA4Seg_ICCV19} &87.3& 35.7& 79.5& 32.0& 14.5& 21.5& 24.8& 13.7& 80.4& 32.0& 70.5& 50.5& 16.9& 81.0& 20.8& 28.1& 4.1& 15.5& 4.1& 37.5  \\
 &ADVENT \cite{vu2018advent}&86.9 &28.7& 78.7& 28.5 &25.2& 17.1& 20.3 &10.9& 80.0& 26.4& 70.2& 47.1& 8.4 &81.5 &26.0& 17.2& \textbf{18.9}& 11.7 &1.6 &36.1\\
 \cline{2-22}
 &Source only&35.4&13.2&72.1&16.7&11.6&20.7&22.5&13.1&76.0&7.6&66.1&41.1&19.0&69.8&15.2&16.3&0.0&16.2&4.7&28.3\\
 & Baseline (feat. only) \cite{Tsai_adaptseg_2018} &85.7 &22.8& 77.6& 24.8& 10.6& 22.2& 19.7& 10.8& 79.7& 27.8& 64.8& 41.5& 18.4& 79.7& 19.9& 21.8& 0.5& 16.2& 4.2& 34.1\\
 &FADA &\textbf{92.3}&\textbf{51.1}&\textbf{83.7}&\textbf{33.1}&\textbf{29.1}&\textbf{28.5}&28.0&\textbf{21.0}&\textbf{82.6}&32.6&\textbf{85.3}&\textbf{55.2}&\textbf{28.8}&\textbf{83.5}&24.4&\textbf{37.4}&0.0&\textbf{21.1}&15.2&\textbf{43.8}\\
 \hline
 
\multirow{9}{*}{ResNet-101}
&AdaptSegNet \cite{Tsai_adaptseg_2018} & 86.5& 36.0& 79.9& 23.4& 23.3& 23.9& 35.2& 14.8& 83.4& 33.3& 75.6& 58.5& 27.6& 73.7& 32.5& 35.4& 3.9& 30.1& 28.1& 42.4\\
&SIBAN \cite{Luo_2019_ICCV}&88.5 &35.4 &79.5& 26.3& 24.3& 28.5& 32.5& 18.3& 81.2& \textbf{40.0}& 76.5& 58.1& 25.8& 82.6& 30.3& 34.4& 3.4& 21.6& 21.5& 42.6\\

&CLAN \cite{Yawei2019Taking} & 87.0 &27.1& 79.6& 27.3& 23.3& 28.3& \textbf{35.5}& 24.2& 83.6& 27.4& 74.2& 58.6& 28.0& 76.2& 33.1& 36.7& 6.7& 31.9& 31.4& 43.2\\
& AdaptPatch \cite{Tsai_DA4Seg_ICCV19} &92.3 &\textbf{51.9}& 82.1& 29.2& 25.1& 24.5& 33.8& \textbf{33.0}& 82.4& 32.8& 82.2& 58.6& 27.2& 84.3& 33.4& 46.3& 2.2& 29.5& 32.3& 46.5  \\
&ADVENT \cite{vu2018advent}&89.4 &33.1 &81.0 &26.6 &26.8 &27.2 &33.5 &24.7 &83.9 &36.7 &78.8 &58.7 &30.5 &84.8 &\textbf{38.5} &44.5 &1.7 &31.6& 32.4 &45.5\\
\cline{2-22}
& Source only &65.0&16.1&68.7&18.6&16.8&21.3&31.4&11.2&83.0&22.0&78.0&54.4&33.8&73.9&12.7&30.7&\textbf{13.7}&28.1&19.7&36.8\\
& Baseline (feat. only) \cite{Tsai_adaptseg_2018} &83.7 &27.6& 75.5& 20.3& 19.9& 27.4& 28.3& 27.4& 79.0& 28.4& 70.1& 55.1& 20.2& 72.9& 22.5& 35.7& 8.3& 20.6& 23.0& 39.3\\
& FADA& \textbf{92.5}&47.5&85.1&37.6&\textbf{32.8}&33.4&33.8&18.4&85.3&37.7&83.5&63.2&39.7&\textbf{87.5}&32.9&\textbf{47.8}&1.6&\textbf{34.9}&\textbf{39.5}&49.2\\


& FADA-MST& 91.0&50.6&\textbf{86.0}&\textbf{43.4}&29.8&\textbf{36.8}&43.4&25.0&\textbf{86.8}&38.3&\textbf{87.4}&\textbf{64.0}&\textbf{38.0}&85.2&31.6&46.1&6.5&25.4&37.1&\textbf{50.1}\\

 \hline
\end{tabular}}
\end{center}

\label{tab:gta5_results}
\end{table}

\subsection{Evaluation Metrics}
The metrics for evaluating our algorithm is consistent with the common semantic segmentation task. Specifically, we compute PSACAL VOC intersection-over-union ($\mathbf{IoU}$) \cite{pascal_voc} of our prediction and the ground truth label. We have $\mathbf{IoU}=\frac{\mathsf{TP}}{\mathsf{TP}+\mathsf{FP}+\mathsf{FN}}$, where $\mathsf{TP}$, $\mathsf{FP}$ and $\mathsf{FN}$ are the numbers of true positive, false positive and false negative pixels respectively. In addition to the $\mathbf{IoU}$ for each class, a $\mathbf{mIoU}$ is also reported as the mean of $\mathbf{IoU}$s over all classes. 

\subsection{Implementation Details}
Our pipeline is implemented by PyTorch \cite{paszke2017pytorch}. For fair comparison, we employ DeeplabV2 \cite{deeplabv2} with VGG-16 \cite{vgg} and ResNet-101 \cite{resnet} as the segmentation base networks. All models are pre-trained on ImageNet \cite{imagenet_cvpr09}. For the fine-grained discriminator, we adopt a simple structure consisting of 3 convolution layers with channel numbers $\{256, 128, 2K\}$, $3\times3$ kernels, and stride of 1. Each convolution layer is followed by a Leaky-ReLU \cite{leaky_relu} parameterized by 0.2 except for the last layer.

To train the segmentation network, we use the Stochastic Gradient
Descent (SGD) optimizer where the momentum is 0.9 and the weight decay is $10^{-4}$. The learning rate is initially set to $2.5\times10^{-4}$ and is decreased following a `poly' learning rate policy with power of 0.9. For training the discriminator, we adopt the Adam optimizer with $\beta_1=0.9$, $\beta_2=0.99$ and the initial learning rate as $10^{-4}$. The same 'poly' learning rate policy is used. $\lambda_{adv}$ is constantly set to 0.001. Temperature T is set as 1.8 for all experiments.

Regarding the training procedure, the network is first trained on source data for 20k iterations and then fine-tuned using our framework for 40k iterations. The batch size is eight. Four are source images and the other four are target images. Some data augmentations are used including random flip and color jittering to prevent overfitting.

Although our model is already able to achieve new state-of-the-art results, we further boost the performance by using self distillation \cite{born_again, Label_Refinery, gift_knowledge_distillation} and multi-scale testing. A detailed ablation study is conducted in Section \ref{sec:analysis} to reveal the effect of each component, which, we hope, could provide more insights into the topic.


\subsection{Comparison with State-of-the-art Methods}
\noindent\textbf{Small shift: Cross city adaptation.}\quad
Adaptation between real images from different cities is a scenario with great potential for practical applications. Table \ref{tab:crosscity_table} shows the results of domain adaptation on Cityscapes $\to$ Cross-City dataset. 
Our method has different performance gains for the four cities. 
On average over four cities, our FADA achieves 8.5\% improvement compared with the source-only baselines, and 2.25\% gain compared with the previous best method.

\begin{figure}
	\centering
	\scalebox{1}{
		\begin{tabular}{cccc}
			\subfloat{\includegraphics[width=0.2\linewidth]{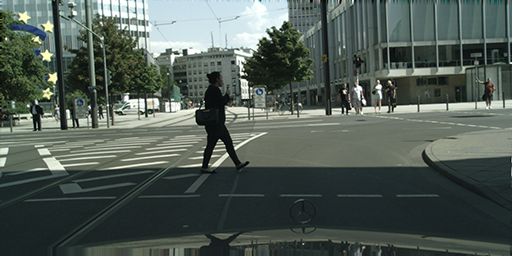}} &
			\subfloat{\includegraphics[width=0.2\linewidth]{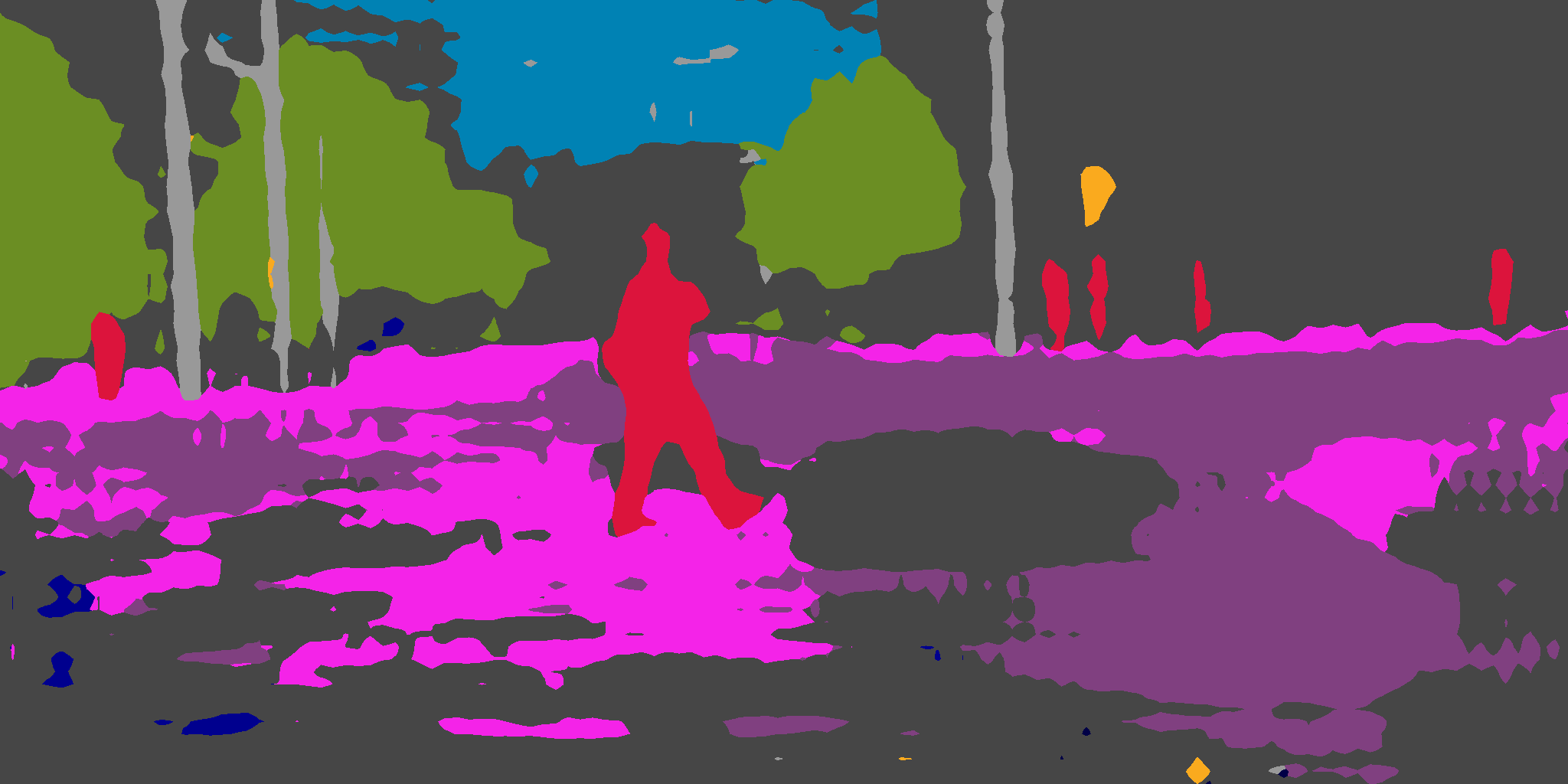}} &
			\subfloat{\includegraphics[width=0.2\linewidth]{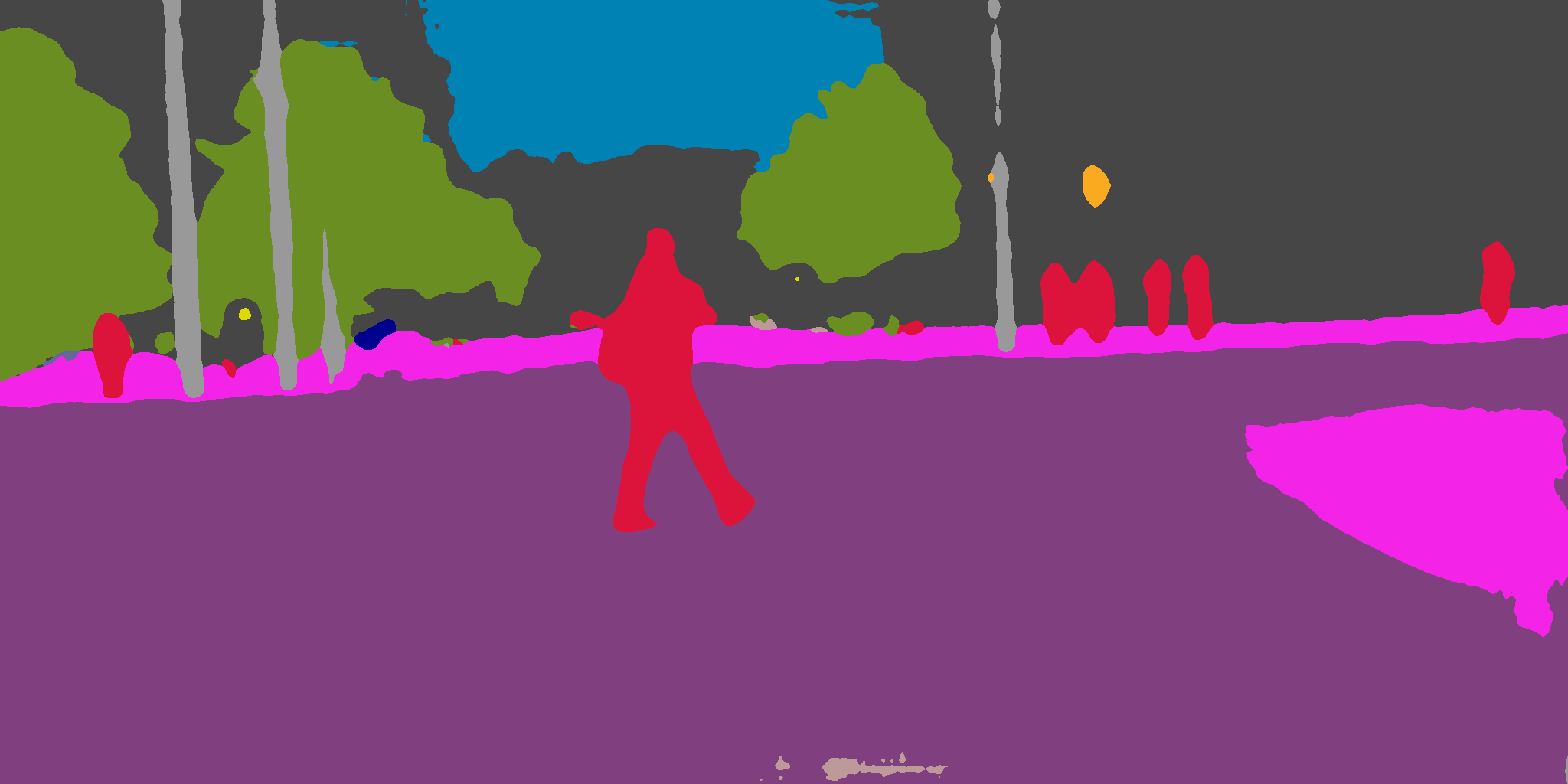}} &
			\subfloat{\includegraphics[width=0.2\linewidth]{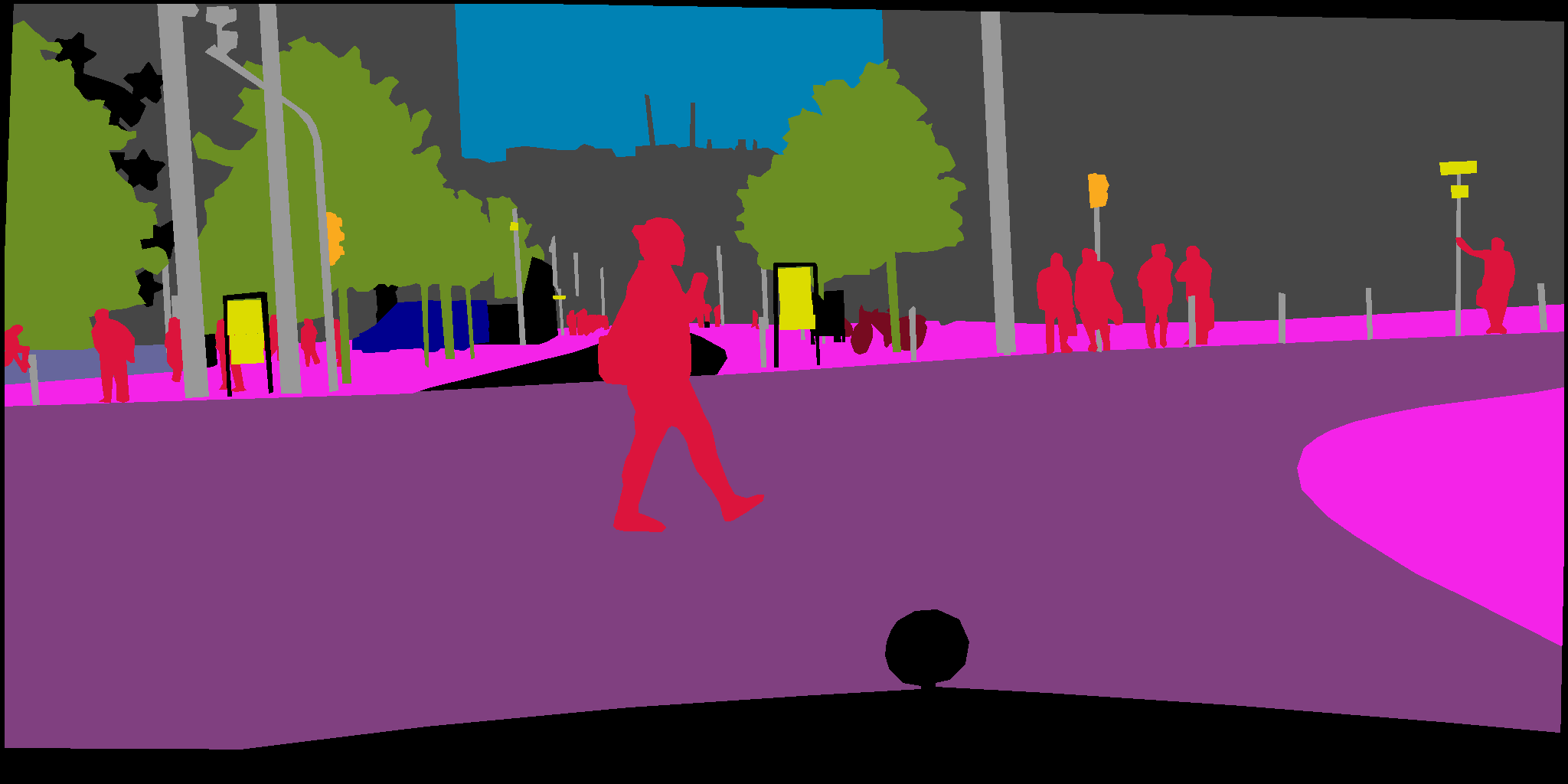}} \\
			
			\subfloat{\includegraphics[width=0.2\linewidth]{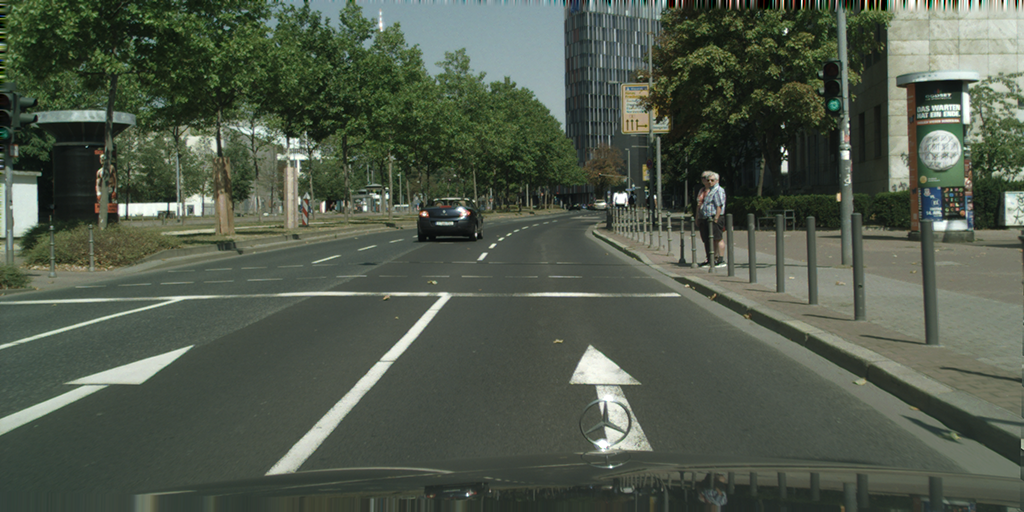}} &
			\subfloat{\includegraphics[width=0.2\linewidth]{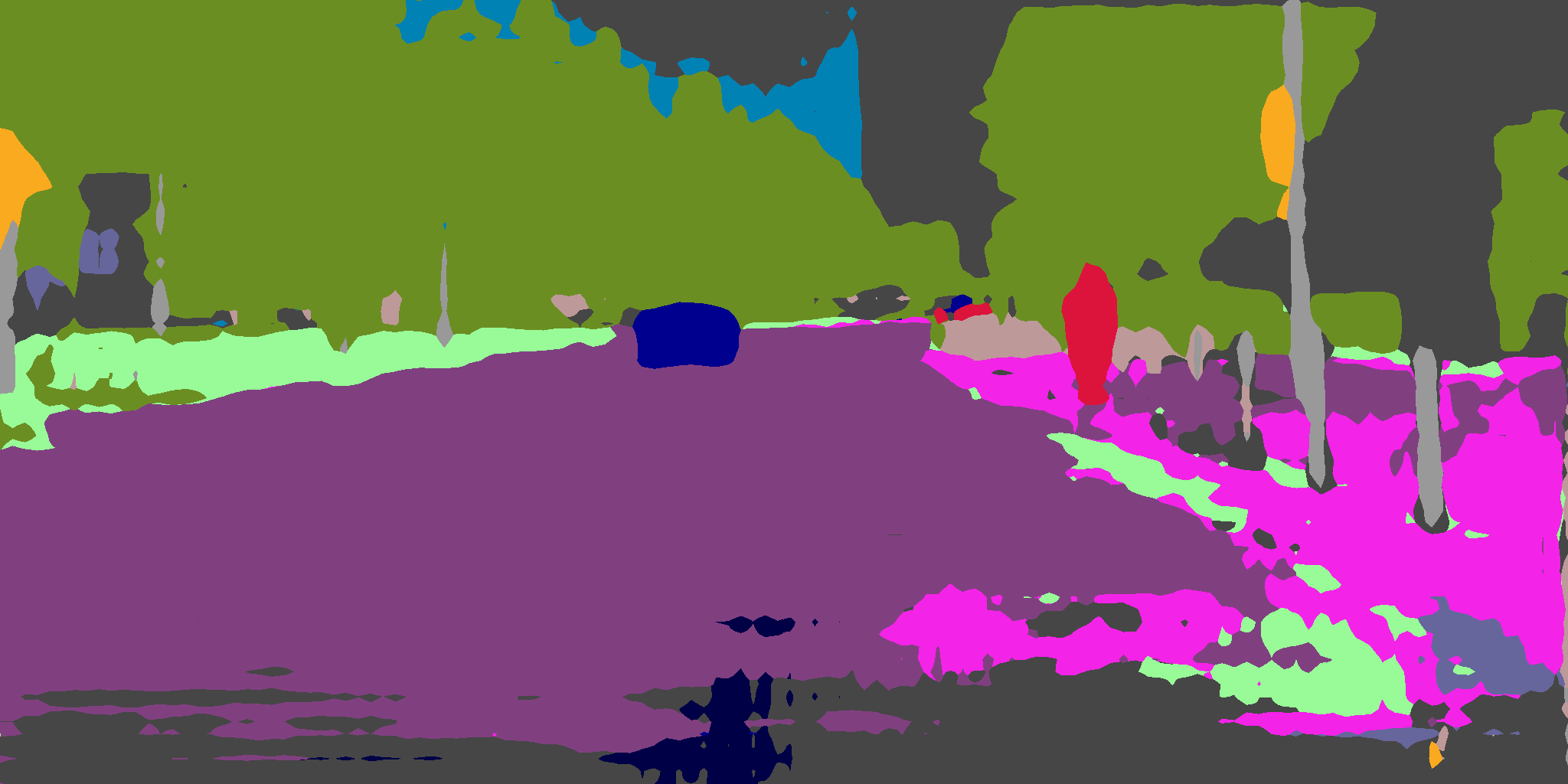}} &
			\subfloat{\includegraphics[width=0.2\linewidth]{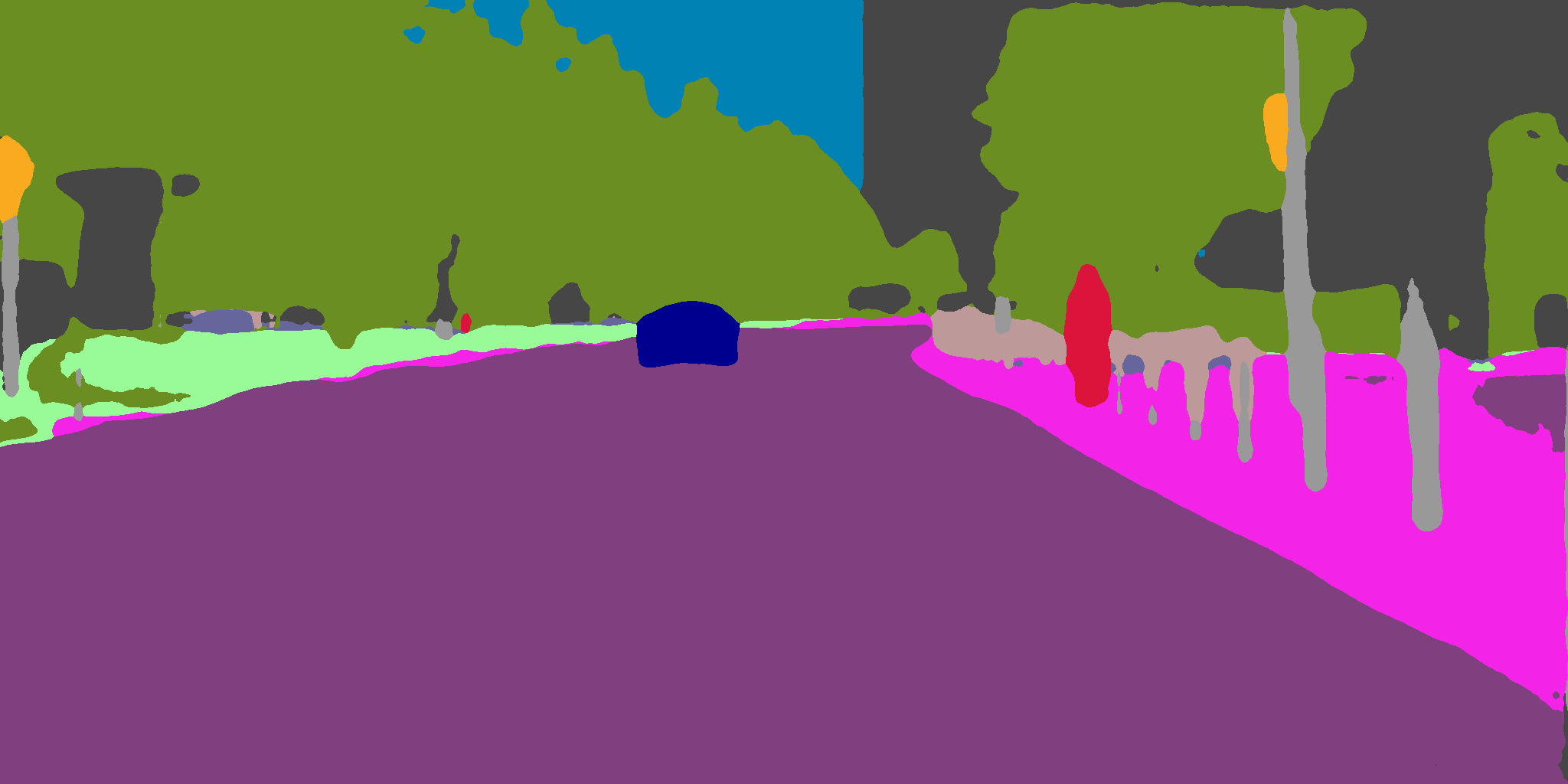}} &
			\subfloat{\includegraphics[width=0.2\linewidth]{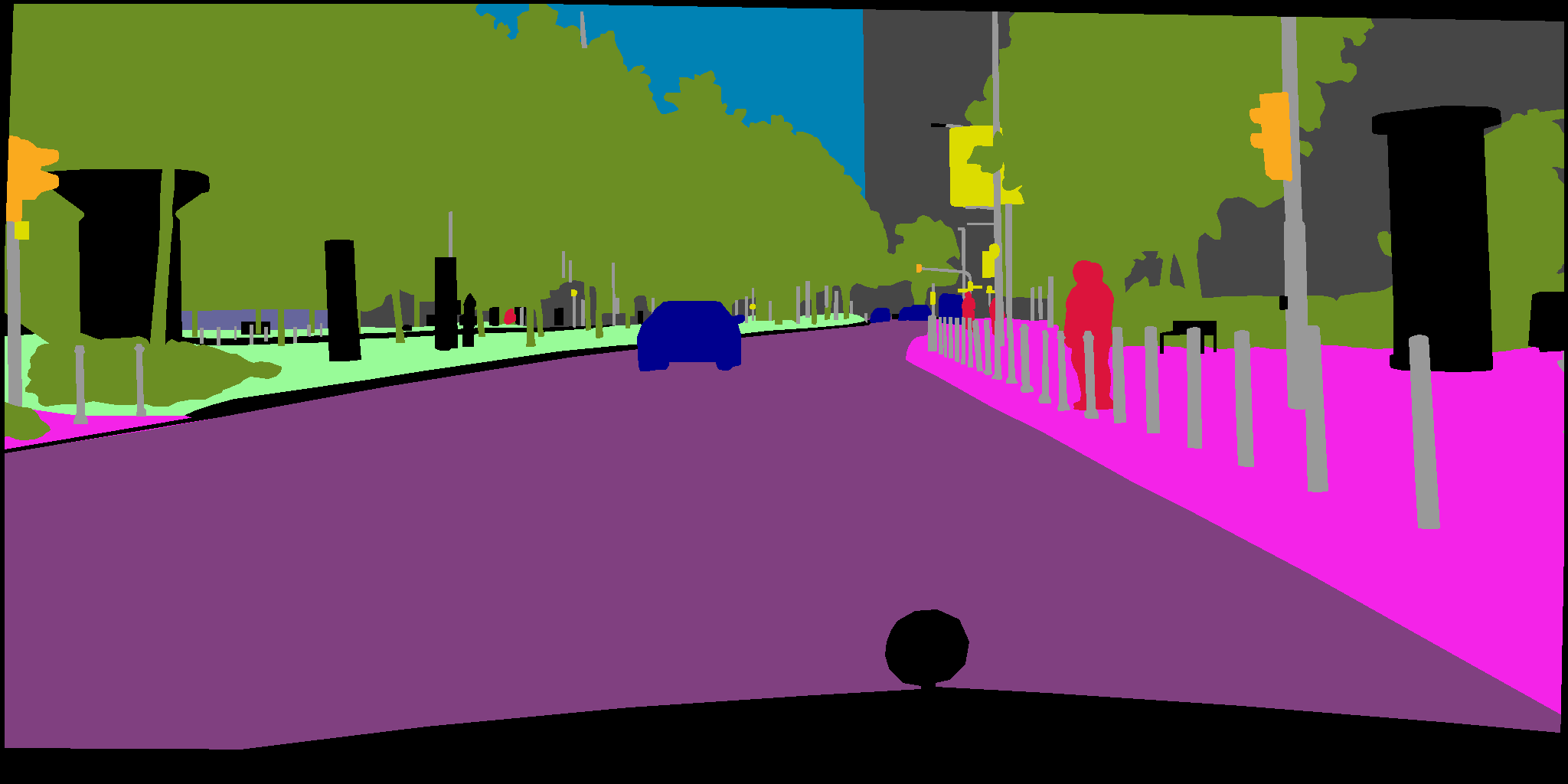}} \\
			
			\subfloat{\includegraphics[width=0.2\linewidth]{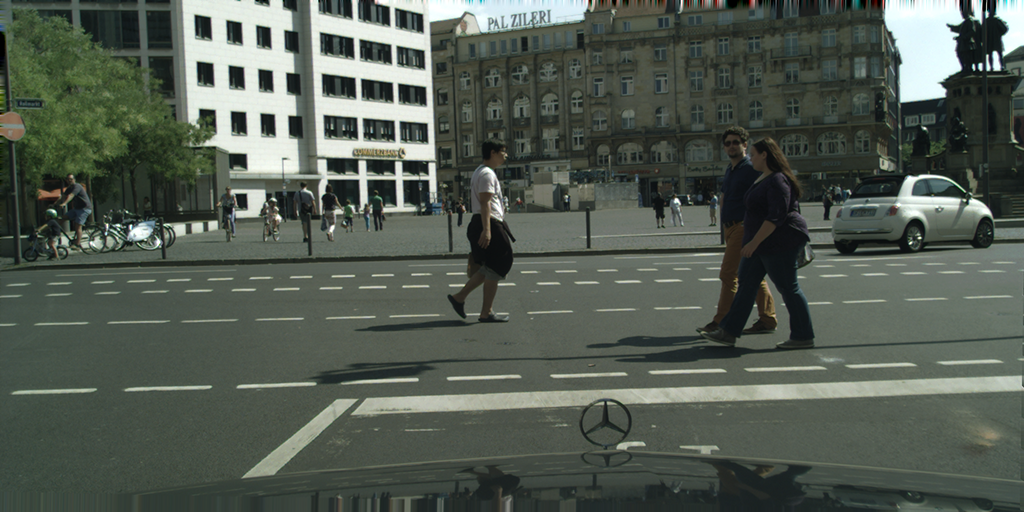}} &
			\subfloat{\includegraphics[width=0.2\linewidth]{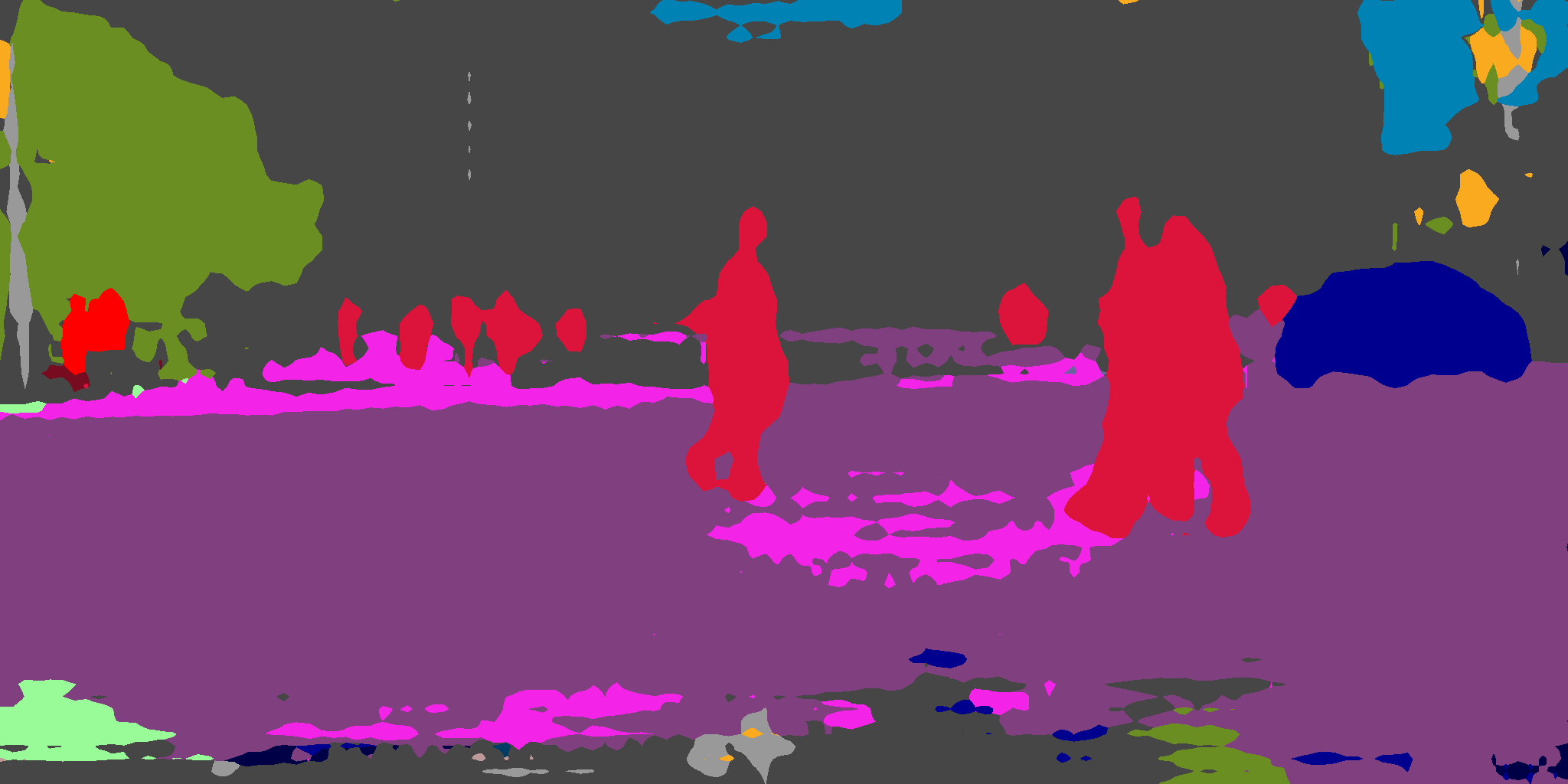}} &
			\subfloat{\includegraphics[width=0.2\linewidth]{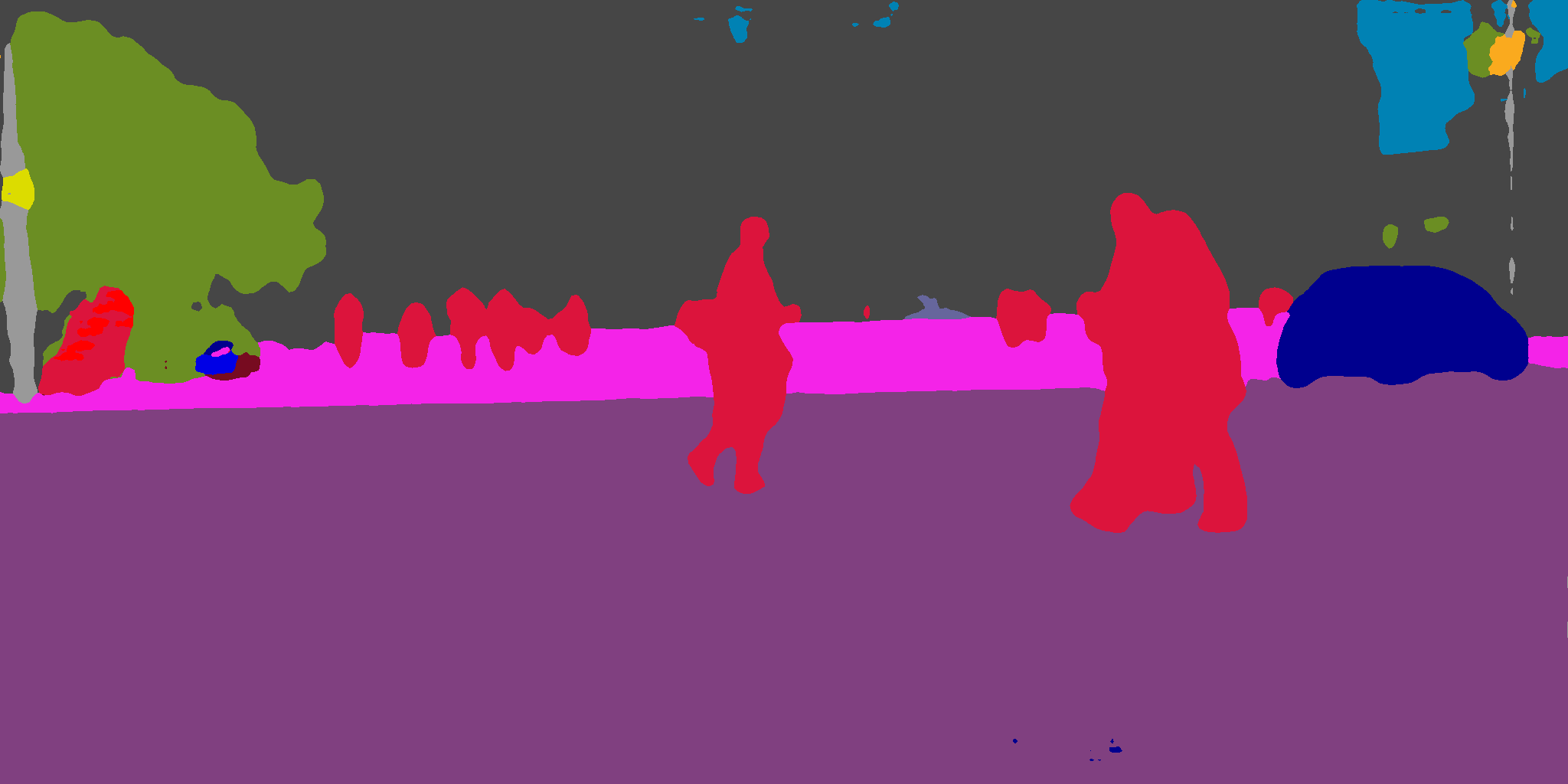}} &
			\subfloat{\includegraphics[width=0.2\linewidth]{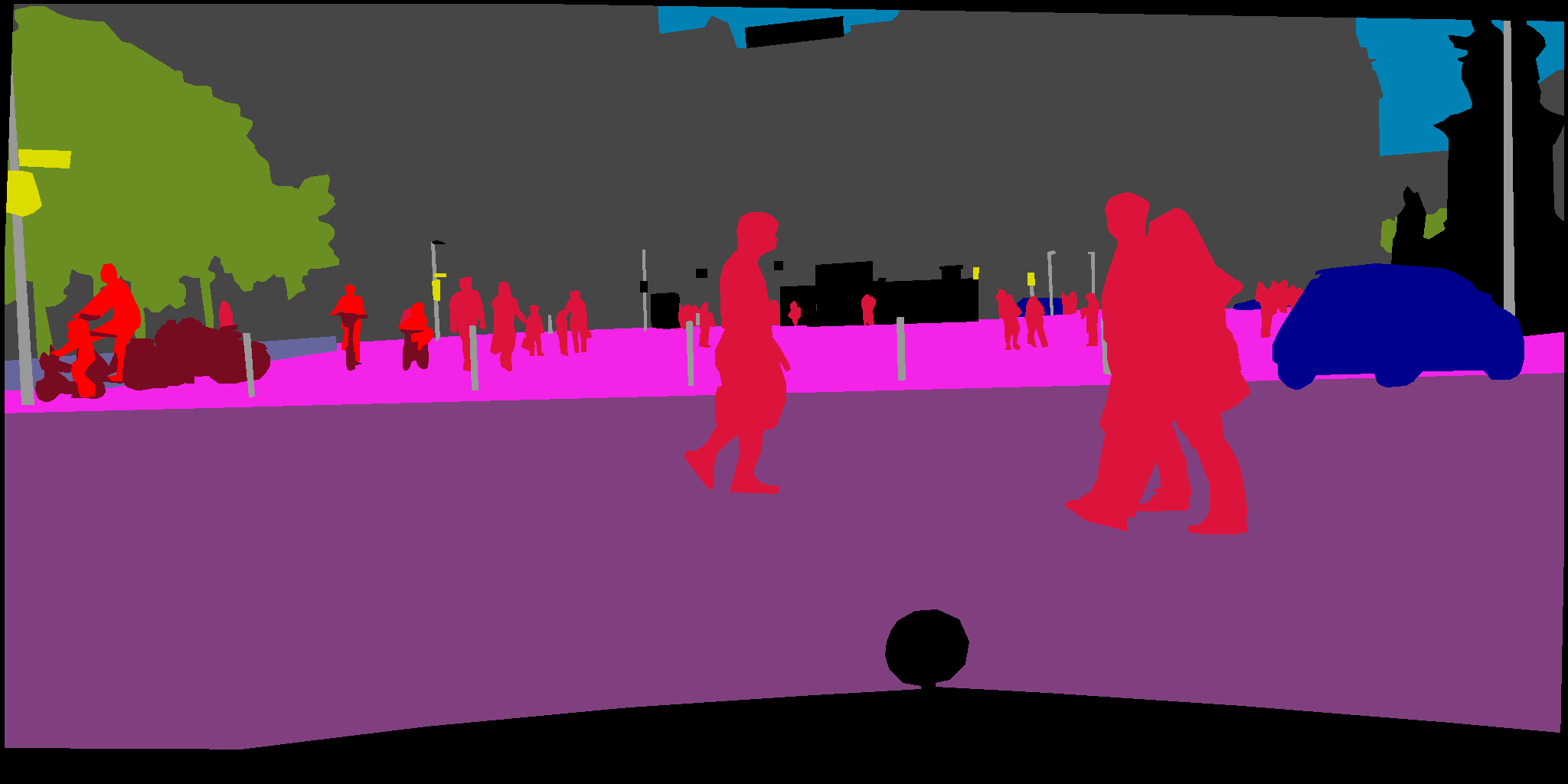}}\\
			
			\subfloat{\includegraphics[width=0.2\linewidth]{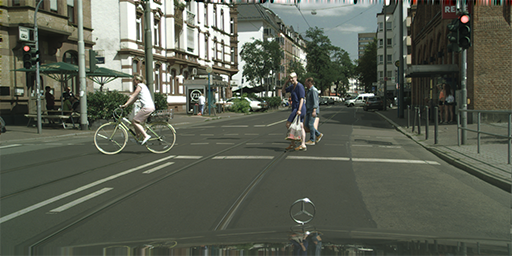}} &
			\subfloat{\includegraphics[width=0.2\linewidth]{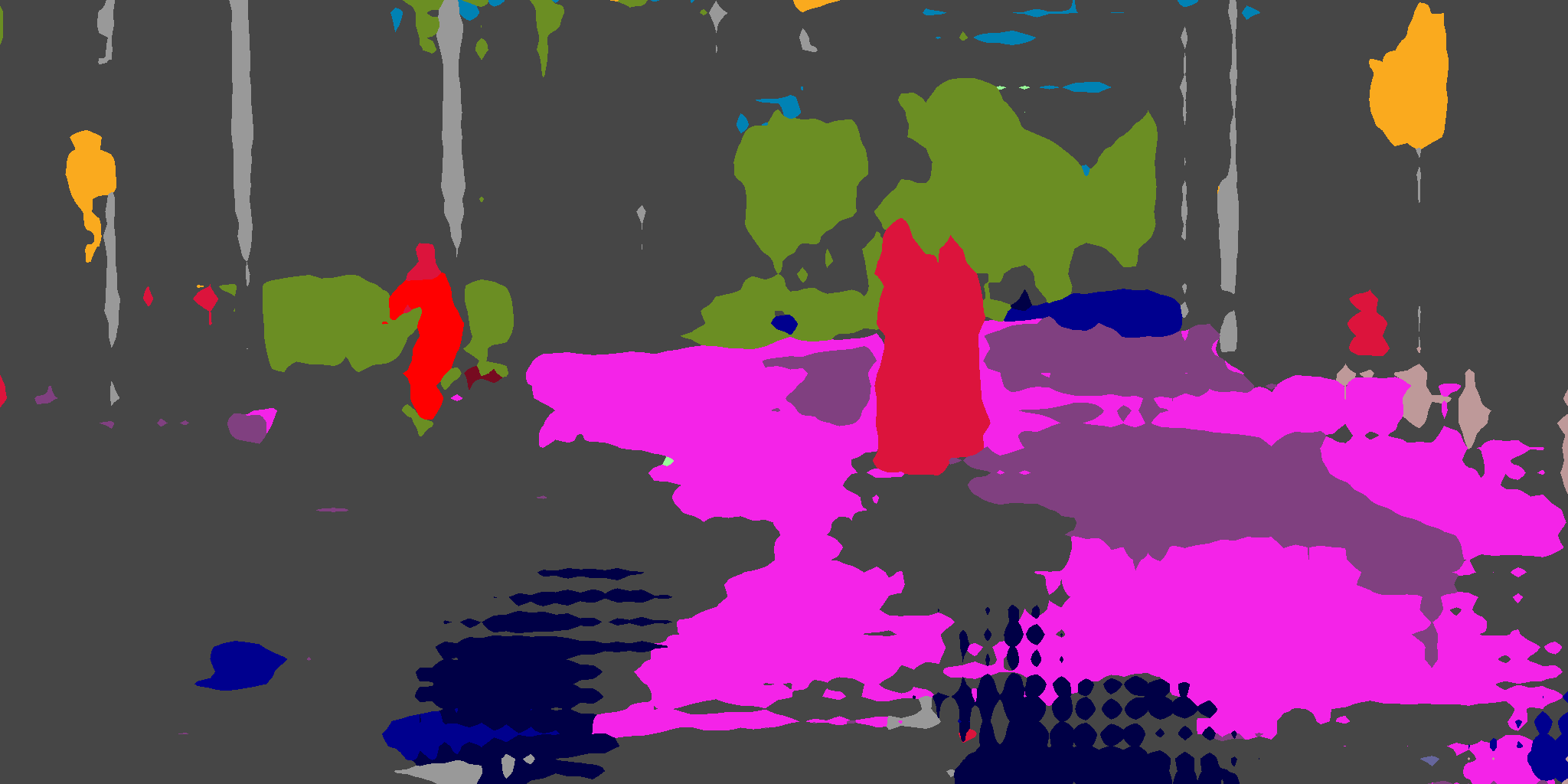}} &
			\subfloat{\includegraphics[width=0.2\linewidth]{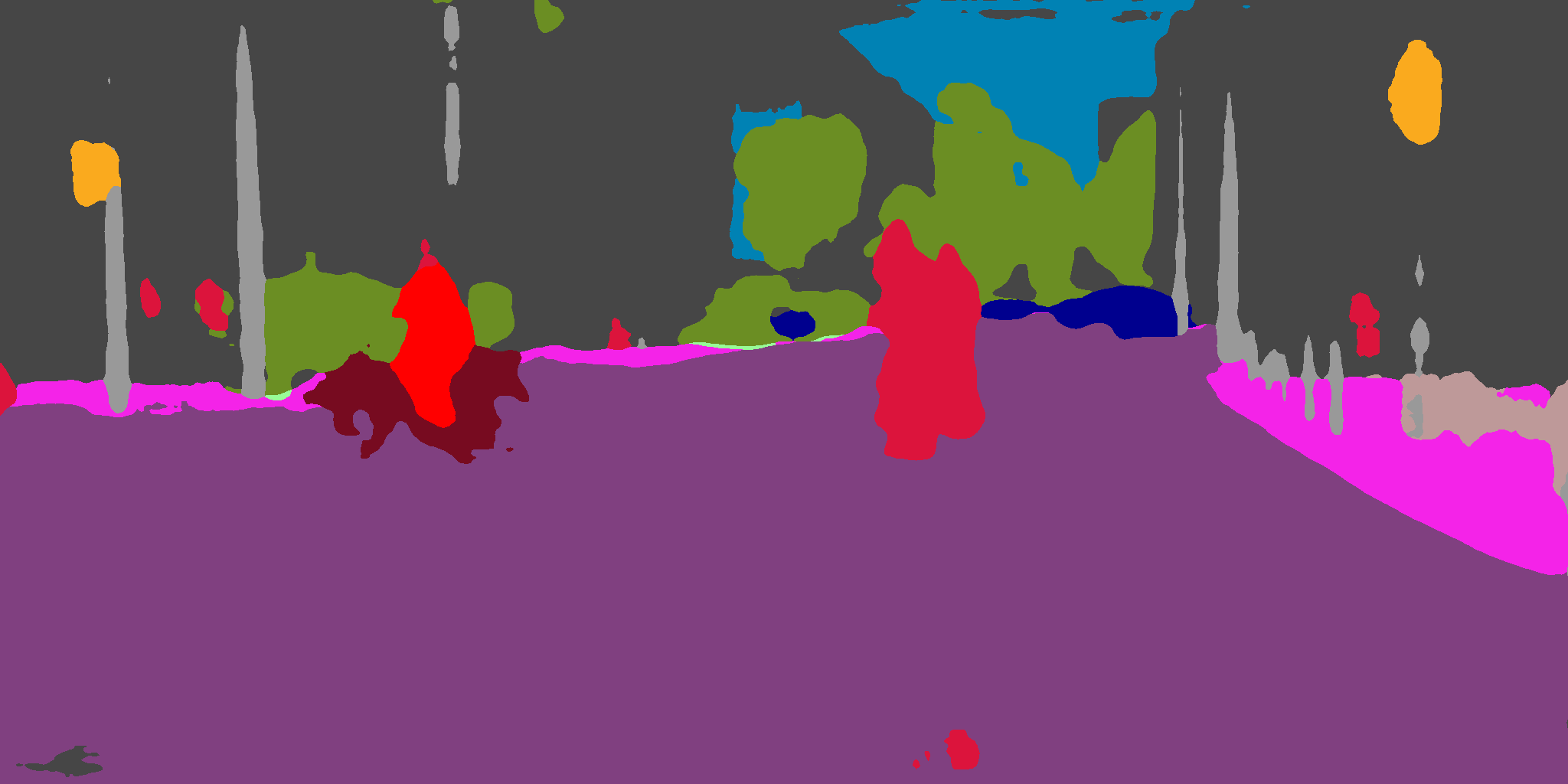}} &
			\subfloat{\includegraphics[width=0.2\linewidth]{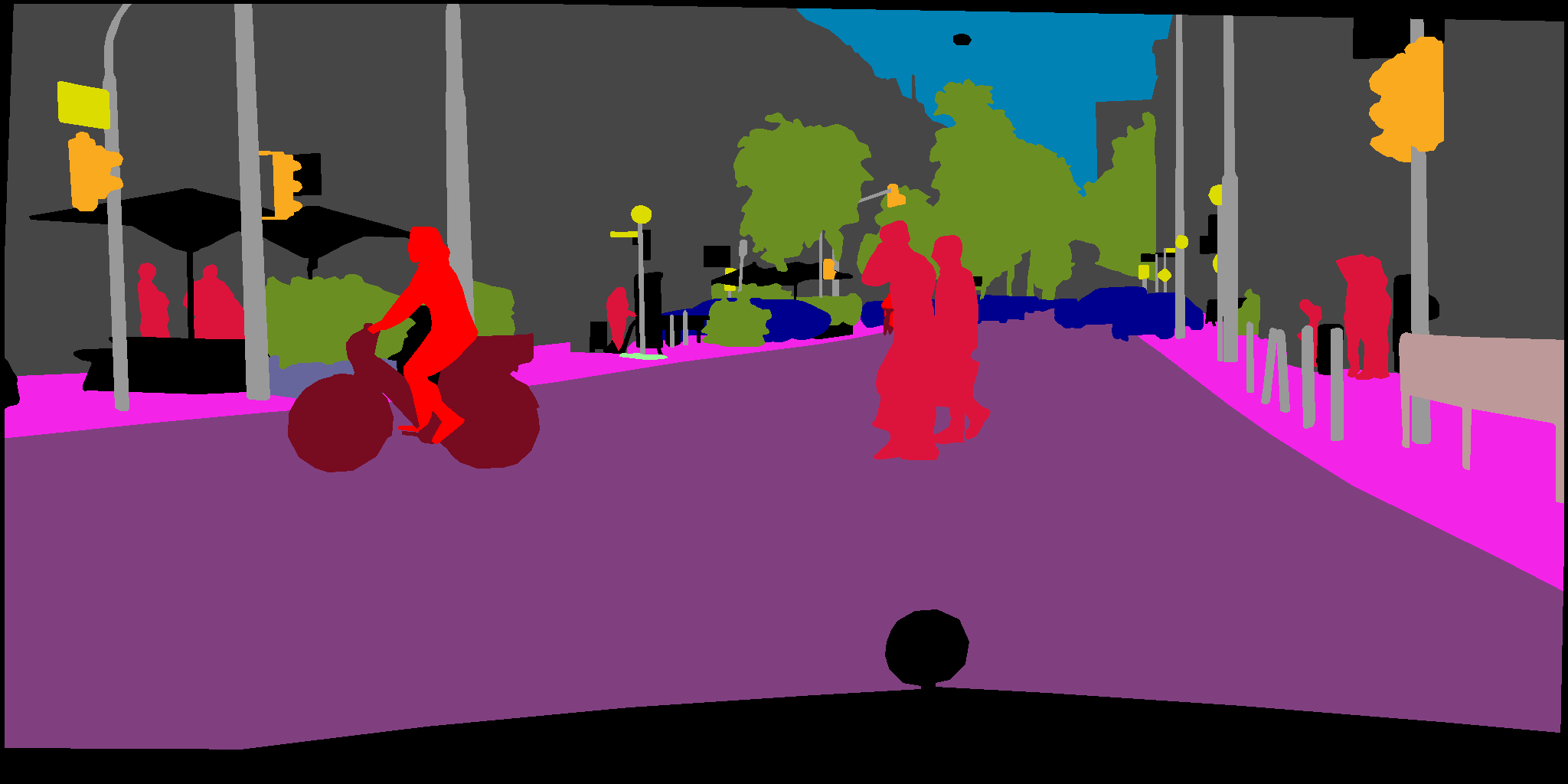}} \\

			Image & Before adaptation & After adaptation & Ground-truth
	\end{tabular}}
	\caption{Qualitative segmentation results for GTA5 $\to$ Cityscapes.}
	\label{fig:gta5_images}
\end{figure}

\noindent\textbf{Large shift: Synthetic to real adaptation.}\quad
Table \ref{tab:synthia_results} and \ref{tab:gta5_results}  demonstrate the semantic segmentation performance on SYNTHIA $\to$ Cityscapes and GTA5 $\to$ Cityscapes tasks in comparison with existing state-of-the-art domain adaptation methods. We could observe that our FADA outperforms the existing methods by a large margin and obtain new state-of-the-art performance in terms of mIoU. Compared to the source model without any adaptation, a gain of 16.4\% and 13.9\% are achieved for VGG16 and ResNet101 respectively on SYNTHIA $\to$ Cityscapes. FADA also obtains 15.5\% and 12.4\% improvement on different baelines for GTA5 $\to$ Cityscapes task. Besides, compared to the state-of-the-art feature-level methods, a general improvement of over 4\% is witnessed. 
Note that as mentioned in \cite{Zhang_2017_ICCV_CDA}, the ``train" images in Cityscapes are more visually similar to the ``bus" in GTA5 instead of the ``train" in GTA5, which is also a challenge to other methods. Qualitative results for GTA5 $\to$ Cityscapes task are presented at Figure \ref{fig:gta5_images}, reflecting that FADA also brings a significant visual improvement.

\subsection{Feature distribution}
\label{sec:analysis}
\begin{figure}[t]
	\begin{center}
		\includegraphics[width=\linewidth]{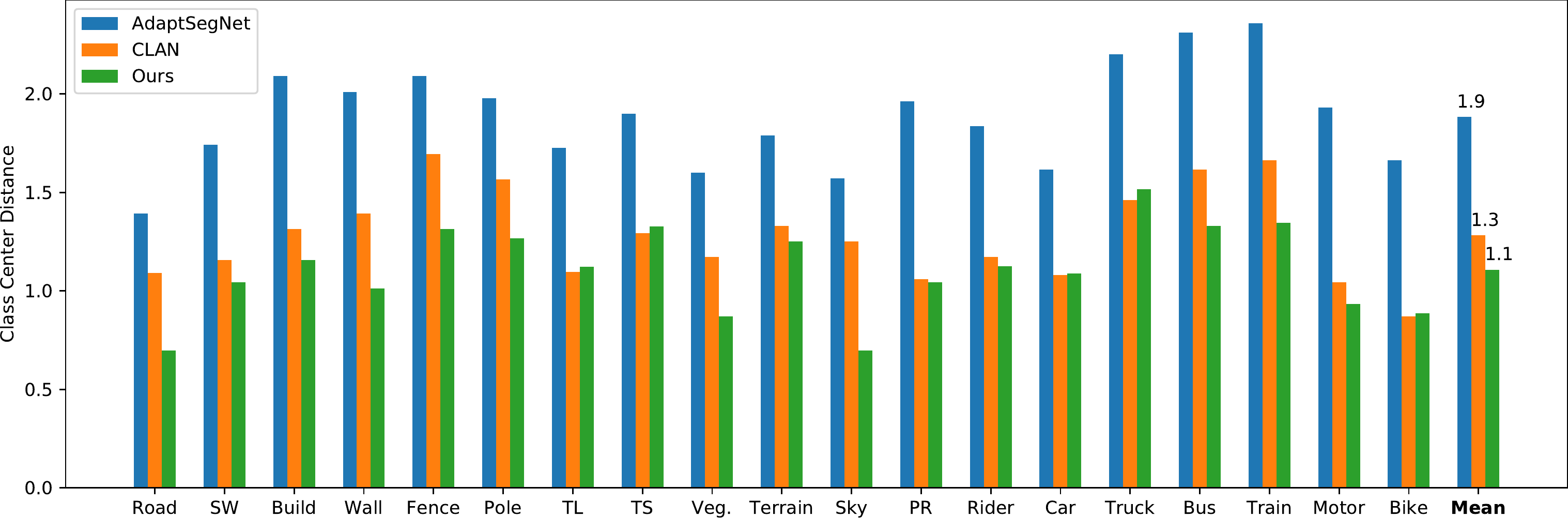}
	\end{center}
	\caption{Quantitative analysis of the feature joint distributions. For each class, we show the Class Center Distance as defined in Equation \ref{CCD}. Our FADA shows a better aligned structure in class-level compared with other state-of-the-art methods.}
	\label{fig:ccd}
\end{figure}
To verify whether our fine-grained adversarial framework aligns features on a class-level, we design an experiment to investigate to what degree the class-level features are aligned. Considering different networks map features to different feature spaces, it's necessarily to find a stable metric. CLAN \cite{Yawei2019Taking} suggests to use a Cluster Center Distance, which is defined as the ratio of intra-class distance between the trained model and the initial model, to measure class-level alignment degree. To better evaluate the effectiveness of class-level feature alignment on the same scale, we propose to modify the Cluster Center Distance to the Class Center Distance (CCD) by taking inter-class distance into account. The CCD for class $i$ is defined as follows:
\begin{equation}\label{CCD}
CCD(i)=\frac{1}{K-1}\sum_{j=1,j\neq i}^{K}\frac{\frac{1}{|S_i|}\sum_{\mathbf{x}\in S_i}{\|\mathbf{x}-\mu_i\|^2}}{\|\mu_i-\mu_j\|^2}
\end{equation}
where $\mu_i$ is the class center for class $i$, $S_i$ is the set of all features belonging to class $i$. With CCD, we could measure the ratio of intra-class compactness over inter-class distance. A low CCD suggests the features of same class are clustered densely while the distance between different classes is relatively large. We randomly pick 2,000 source samples and 2,000 target samples respectively, and compare the CCD values with other state-of-the-art methods: AdaptSegNet for global alignment and CLAN for class-wise alignment without explicitly modeling the class relationship. As shown in the Figure \ref{fig:ccd}, FADA achieves a much lower CCD on most classes and get the lowest mean CCD value 1.1 compared to other algorithms. With FADA, we can achieve better class-level alignment and preserve consistent class structures between domains.

\subsection{Ablation studies}
\noindent\textbf{Analysis of different components.}\quad
Table \ref{tab:diff_comp} presents the impact of each component on DeeplabV2 with ResNet-101 on GTA5 $\to$ Cityscapes task. The fine-grained adversarial training brings an improvement of 10.1\%, which already makes it the new state of the art. To further explore the potential of the model, the self distillation strategy leads to an improvement of 2.3\% and multi-scale testing further boosts the performance by 0.7\%.

\noindent\textbf{Hard labels vs. Soft labels.}
\quad
\label{vs}
As discussed in Section \ref{class_level_info}, the knowledge extracted from the classifier C could be produced from hard labels or soft labels. Here we compare these two forms of label on GTA5 $\to$ Cityscapes and SYNTHIA $\to$ Cityscapes tasks with DeeplabV2 ResNet-101. For soft labels, we use ''confidence clipping`` with threhold 0.9 as regularization. For hard labels, we only keep high-confidence samples, while ignoring the samples with confidence lower than 0.9. The results are reported in Table \ref{tab:hard_vs_soft}. Both choices give great boost to the baseline global feature alignment model. We observe that soft label is a more flexible choice and present more superior performance.


\noindent\textbf{Impact of Confidence Clipping.}\quad
In our experiments, we use "confidence clipping" as a regularizer to prevent overfitting on noisy soft labels. The values of the confidence are truncated by a given threshold, therefore the values are not encouraged to heavily fit to a certain class. We test several thresholds and the results are shown in Table \ref{tab:confidence_clipping}. Note that when the threshold is 1.0, it means no regularization is used. We observe constant performance gain using the confidence clipping. The best result is found when the threshold is 0.9.

\begin{table}[t]
    \caption{Ablation studies of each component. F-Adv refers to fine-grained adversarial training; SD refers to self distillation; MST refers to multi-scale testing.}
    \centering
    \begin{tabular}{c c c c }
    \hline
         F-Adv  & SD &MST &mIoU  \\
         \hline
         &  & &36.8 \\
         \checkmark&&&46.9\\
         \checkmark&\checkmark&&49.2\\
         \checkmark&\checkmark&\checkmark&\textbf{50.1}\\
         \hline

    \end{tabular}
    
    \label{tab:diff_comp}
\end{table}

\begin{table}
    \caption{Comparison of different strategies for extracting class-level knowledge on GTA5 $\to$ Cityscapes and SYNTHIA $\to$ Cityscapes tasks.}
    \centering
    \begin{tabular}{c c c}\hline
     &GTA5&SYNTHIA\\
    \hline
    baseline \cite{Tsai_adaptseg_2018}     &  39.4 & 35.4\\
    hard labels & 45.7 & 40.8\\
    soft labels   &  \textbf{46.9} & \textbf{41.5}  \\\hline
    \end{tabular}
    \label{tab:hard_vs_soft}
\end{table}


\begin{table}
    \caption{Influence of threshold for confidence clipping.}
    \centering
    \label{tab:confidence_clipping}
    \begin{tabular}{c  c c c c}
    \hline
    \multicolumn{5}{c}{\textbf{GTA5} $\to$ \textbf{Cityscapes}}\\
    \hline
    threshold & 0.7  & 0.8  & 0.9  & 1.0 \\
         mIoU & 46.2 & 46.3 & \textbf{46.9} & 45.7  \\
         \hline
    \end{tabular}
    
\end{table}

\section{Conclusion}
In this paper, we address the problem of domain adaptive semantic segmentation by proposing a fine-grained adversarial training framework. A novel fine-grained discriminator is designed to not only distinguish domains, but also capture category-level information to guide a fine-grained feature alignment. The binary domain labels used to supervise the discriminator are generalized to domain encodings correspondingly to incorporate class information. 
Comprehensive experiments and analysis validate the effectiveness of our method. Our method achieves new state-of-the-art results on three popular tasks, outperforming other methods by a large margin.

\noindent\textbf{Acknowledgement:} This work was partially supported by Beijing Academy of Artificial Intelligence (BAAI).



\clearpage
%
%
\bibliographystyle{splncs04}
\bibliography{egbib}
\end{document}